\documentclass[10pt,twocolumn,letterpaper]{article}

\usepackage{iccv}
\usepackage{times}
\usepackage{epsfig}
\usepackage{graphicx}
\usepackage{amsmath}
\usepackage{booktabs}
\usepackage{amssymb}
\usepackage[T1]{fontenc}
\usepackage[utf8]{inputenc}
\usepackage{authblk}
\usepackage[breaklinks=true,bookmarks=false]{hyperref}
\usepackage[capitalize]{cleveref}
\crefname{section}{Sec.}{Secs.}
\Crefname{section}{Section}{Sections}
\Crefname{table}{Table}{Tables}
\crefname{table}{Tab.}{Tabs.}
\usepackage[ruled,linesnumbered]{algorithm2e}
\usepackage[accsupp]{axessibility}
\usepackage{multicol}
\usepackage{multirow}
\usepackage[table]{xcolor}

\iccvfinalcopy 

\def\iccvPaperID{****} 

\ificcvfinal\fi
\begin{document}

\title{Separable Multi-Concept Erasure from Diffusion Models}

\author{Mengnan Zhao}
\author{Lihe Zhang}
\author{Tianhang Zheng}
\author{Yuqiu Kong}
\author{Baocai Yin}
\affil{\url{https://github.com/Dlut-lab-zmn/SepCE4MU}}

\renewcommand\Authands{ and }
\maketitle
\ificcvfinal\thispagestyle{empty}\fi

\begin{abstract}
Large-scale diffusion models, known for their impressive image generation capabilities, have raised concerns among researchers regarding social impacts, such as the imitation of copyrighted artistic styles. 
In response, existing approaches turn to machine unlearning techniques to eliminate unsafe concepts from pre-trained models. 
However, these methods compromise the generative performance and neglect the coupling among multi-concept erasures, as well as the concept restoration problem.
To address these issues, we propose a Separable Multi-concept Eraser (SepME), which mainly includes two parts: the generation of concept-irrelevant representations and the weight decoupling. 
The former aims to avoid unlearning substantial information that is irrelevant to forgotten concepts.
The latter separates optimizable model weights, making each weight increment correspond to a specific concept erasure without affecting generative performance on other concepts. 
Specifically, the weight increment for erasing a specified concept is formulated as a linear combination of solutions calculated based on other known undesirable concepts.
Extensive experiments indicate the efficacy of our approach in eliminating concepts, preserving model performance, and offering flexibility in the erasure or recovery of various concepts.
\end{abstract}


\section{Introduction}
\label{sec:intro}
The field of text-to-image generation has witnessed remarkable development \cite{zhou2022towards,ge2023expressive,kim2023dense,ramesh2021zero}, especially the occurrence of diffusion models (DMs) like DALL-E2 \cite{ramesh2022hierarchical} and Stable Diffusion \cite{rombach2022high}. 
As the integration of DMs into practical applications \cite{yang2023diffusion,kawar2023imagic,kazerouni2022diffusion} proves advantageous, 
addressing challenges related to their societal impact increasingly attracts the attention of researchers \cite{chourasia2023forget,huang2023receler,golatkar2023training,li2023self}. 
One crucial challenge arises from diverse training data sources, potentially leading to unsafe image generation \cite{zhang2023forget,gandikota2023erasing},  such as the creation of violent content or the imitation of specific artist styles.
To resolve this concern, the machine unlearning (MU) technique has been proposed \cite{zhang2023generate,kumari2023ablating,song2020denoising,schramowski2023safe}, which involves erasing the impact of specific data points or concepts to enhance model security, without necessitating complete retraining from scratch.

The recent MU research such as Erased Stable Diffusion (ESD) \cite{gandikota2023erasing}, Forget-me-not (FMN) \cite{zhang2023forget}, Safe self-distillation diffusion (SDD) \cite{kim2023towards}, and Ablation Concept (AbConcept) \cite{kumari2023ablating}, can be broadly categorized into untargeted concept erasure ({\em e.g.,}, FMN) and targeted concept erasure ({\em e.g.,}, ESD, SDD and AbConcept). 
Specifically, FMN minimizes the attention maps of forgotten concepts.
In contrast, ESD, SDD, and AbConcept align the denoising distribution of forgotten concepts with a predefined distribution.

Despite recent advancements in MU \cite{hong2023all,ni2023degeneration}, there 
exist several drawbacks.
Firstly, prior efforts concentrate on concept erasure, leading to considerable performance degradation in generative capability. Secondly, current erasure procedures are confined to single-concept elimination and pose challenges when extending them to multi-concept erasure. 
The multi-concept erasure can take two forms: simultaneous erasure of multiple concepts and iterative erasure of multiple concepts. The former means that multiple forgotten concepts are known in advance, while the latter implies that each erasure step only possesses knowledge of its previously forgotten concepts. Lastly, to the best of our knowledge, the concept restoration issue has not been considered. For instance, after the erasure of multiple artistic styles, the model owner may regain the copyright associated with some erased styles.

To address these issues, we propose an innovative Separable Multi-concept Eraser (SepME) that contains the generation of concept-irrelevant representations (G-CiRs) and the weight decoupling (WD). 
Specifically, G-CiRs aims to preserve overall model performance while effectively erasing undesirable concepts $c_f$ through early stopping and weight regularization. Early stopping prevents the unlearning of substantial information irrelevant to $c_f$ when 
$\Delta_\epsilon^{\theta_\text{unlearn}}$ and $\Delta_\epsilon^{\theta_\text{ori}}$
become irrelevant.
Here, we define the difference of noise $\epsilon$ predicted by DMs with and without concept $c_f$ as the representations of $c_f$, {\em i.e.,} $\Delta_\epsilon^{\theta_*}$. ${\theta}_\text{ori}$ and ${\theta}_\text{unlearn}$ are weights of the original and unlearned DMs, respectively.
The regularization term restricts the deviation of ${\theta}_\text{unlearn}$ from ${\theta}_\text{ori}$.



Considering the multi-concept erasure and subsequent restoration, WD characterizes the weight variation as ${\bf \Delta}\theta_{{1\sim \text{N}}}$. Each independent weight increment ${\bf \Delta}\theta_{{i}}$ is crafted to erase a specific concept $c_{i,f}$ without compromising the generation performance of models regarding other concepts. More precisely, the weight increment for erasing a specified concept is expressed as a linear combination of particular solutions calculated based on other known undesirable concepts.
Each weight increment shares the same non-zero positions but has distinct values.
These values are determined by the pre-calculated particular solutions and optimizable  linear combination weights.

Our main contributions are summarized as follows:
{\bf (1)} To the best of our knowledge, the scenarios of multi-concept erasure and concept restoration have not been explored in previous literature. This work fills in these critical gaps and designs a separable multi-concept eraser;
{\bf (2)} To effectively unlearn undesirable concepts while maintaining overall model performance, our framework characterizes concept-irrelevant representations;
{\bf (3)} Through extensive experiments, we demonstrate that our method can improve erasing performance, maintain model generation capabilities, and offer flexibility in combining various forgotten concepts, encompassing both deletion and recovery.

\section{Related Works}
The image generation field has experienced rapid development in recent years, evolving from autoencoder \cite{park2020swapping,liu2021variational,shao2020controlvae}, generative adversarial networks \cite{ding2020ccgan,liu2021divco,li2020lightweight}, unconditional diffusion models (DMs) \cite{ho2020denoising,dhariwal2021diffusion} to DMs enhanced with large-scale pre-trained image-text models \cite{gu2022vector,xu2023versatile,kim2022diffusionclip} like CLIP \cite{radford2021learning}. 
These text-guided DMs, exemplified by DALL-E 2 \cite{ramesh2022hierarchical} and Stable Diffusion \cite{rombach2022high}, exhibit an excellent generative ability across various prompts $c$.
The constraint for training DMs is formulated as
\begin{equation*}
    \mathcal{L}_\text{DM} = \mathbb{E}_{x_t\in \mathcal{D}, c, t, \epsilon_\text{GT}\in \mathcal{N}(0,1)} \left[\|\epsilon_\text{GT} - \epsilon_{\theta_\text{dm}}(x_t, c, t)\|_2^2\right],
\end{equation*}
where $x_t$ represents the noised data or the noised latent representation \cite{kingma2013auto}, $x_t = \sqrt{\Bar{\alpha_t}}x_0 + \sqrt{1-\Bar{\alpha_t}}\epsilon$. $\Bar{\alpha_t}$ is the noise variance schedule. $x_0$ denotes the original reference image, and $\epsilon\in \mathcal{N}(0,\mathbf{I})$.
$\mathcal{D}$ is the training dataset. 
$\epsilon_\text{GT}$ means the ground truth noise. $\epsilon_{\theta_\text{dm}}(x_t, c, t)$ denotes the $t$-th step noise predicted by DMs.
$\|\cdot\|_2^2$ is the squared $\ell_2$-norm function.
Additionally, researchers have indicated the unknown concept generation capability of DMs through the fine-tuning of partial model weights on small reference sets \cite{ruiz2023dreambooth,hu2021lora,xu2023qa,gal2022image}. 
Nevertheless, DMs also induce potential risks associated with privacy violations and copyright infringement, such as the training data leakage \cite{carlini2023extracting,duan2023diffusion,dubinski2024towards}, the imitation of various artistic styles \cite{somepalli2023diffusion,shan2023glaze}, and the generation of sensitive content \cite{yang2023mma}. 
Consequently, there is a growing focus on erasing specific outputs from pre-trained DMs \cite{chourasia2023forget,kumari2023ablating}.

Existing research primarily falls into three distinct directions: removal of unsafe data and model retraining \cite{croitoru2023diffusion}, integration of additional plug-ins to guide model outputs \cite{chin2023prompting4debugging,mehrabi2023flirt}, and fine-tuning of model weights through MU techniques \cite{gandikota2023erasing,zhang2023forget,kumari2023ablating}. The drawback of the first direction is that large-scale model retraining demands considerable computational resource and time. The risk of the second  direction is that, with the public availability of model structures and weights, malicious users can easily remove plug-ins. 
This work focuses on the third direction, {\em i.e.,} machine unlearning.

\begin{figure*}[t]
    \begin{center}
        \includegraphics[width=0.9\linewidth]{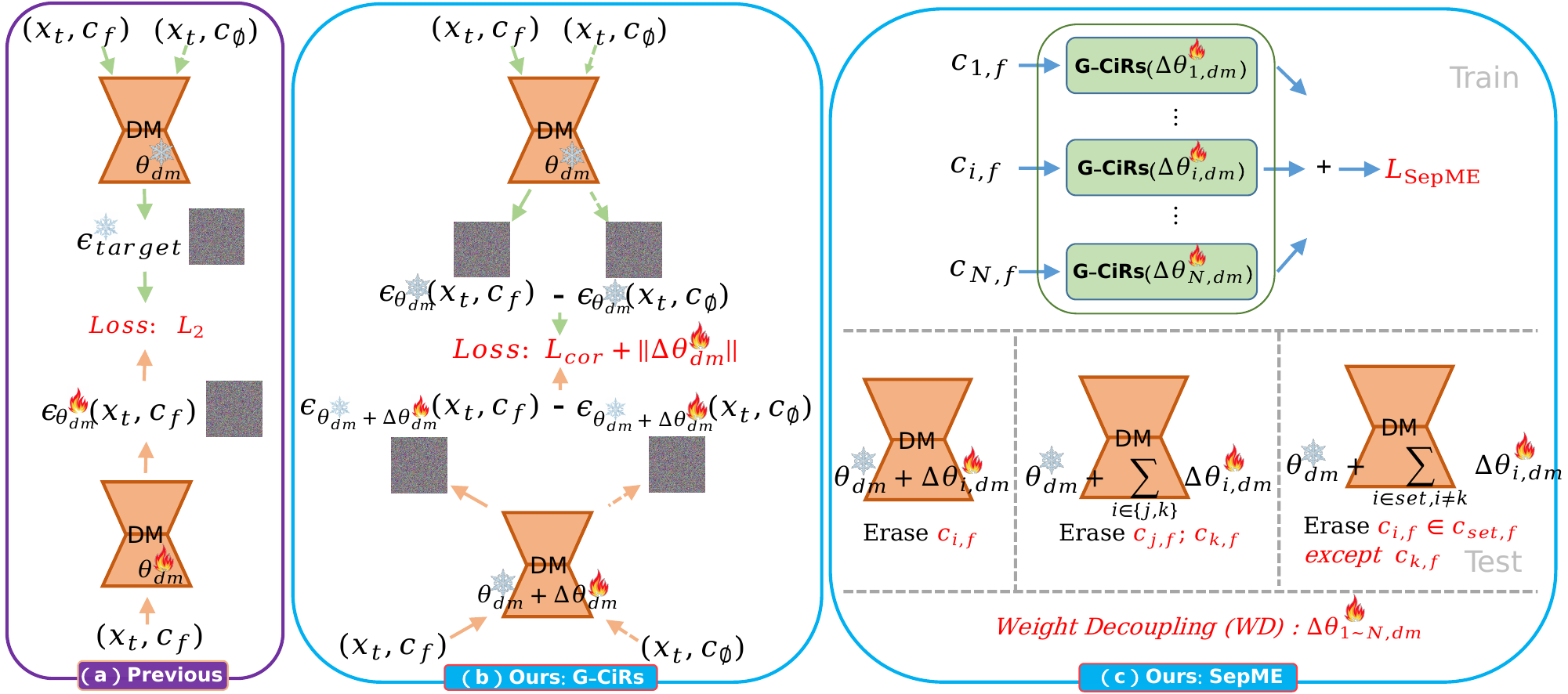}
    \end{center}
    \vspace{-4mm}
    \caption{    
    Overview of various unlearning techniques for DMs.
    `Ice flowers' and `flames' represent frozen and optimizable model weights respectively.
    $\epsilon_\text{target}$ is the predefined noise distribution.
    $\mathcal{L}_2$ denotes the $\ell_2$-norm function.
    $x_t$ means noised samples.
    $c_f$ and $c_\emptyset$ signify the forgotten and blank prompts, respectively.
    $\mathcal{L}_{cor}$ is the correlation function.
    SepME separates optimizable weights as ${\bf \Delta}\theta_{{1\sim \text{N}, \text{dm}}}$, Each ${\bf \Delta}\theta_{{i,\text{dm}}}$ is designed to erase a specific concept $c_{i,f}$ without compromising the generation performance of models regarding other concepts.
    }
    \label{fig1}
    \vspace{-4mm}
\end{figure*}

The majority of unlearning methods for DMs can be summarized as:
\begin{equation}\label{eq4}
	\epsilon_{\theta_\text{op}}(x_t, c_f, t) \leftarrow 
	\begin{cases}
		\epsilon_\text{target} & \text{if } x_0\in \mathcal{D}_{f}\\
		\epsilon_\text{GT} & \text{otherwise},
	\end{cases}
\end{equation}
where $\theta_\text{op}$ represents optimizable model weights, {\em e.g.,} the parameters of cross-attention modules in DMs.
$x_t$ can be obtained through either the diffusion process or the sampling process.
$\epsilon_{\theta_{\text{op}}}(x_t, c_f, t)$ denotes the noise predicted by unlearned DMs at the $t$-th step. $\mathcal{D}_{f}$ refers to the dataset containing the forgotten concept $c_f$.
$\epsilon_{\text{target}}$ and $\epsilon_{\text{GT}}$ represent the noise of predefined target concepts and the ground-truth noise added in the diffusion process, respectively.

For instance, ESD \cite{gandikota2023erasing} leverages the predicted noise for both concept-free $c_\emptyset$ and forgotten concepts $c_f$ to construct $\epsilon_\text{target}$,
\begin{equation*}\label{eq3}
    \epsilon_\text{target} = (1+\eta)\epsilon_{\theta_\text{dm}}(x_t, c_\emptyset, t) - \eta\epsilon_{\theta_\text{dm}}(x_t, c_f, t),
\end{equation*}
where $\theta_\text{dm}$ represents parameters of the frozen DMs.
$\eta$ is the hyperparameter.
SDD \cite{kim2023towards} directly maps the prediction distribution of erased concepts $c_f$ to the prediction distribution of concept-free $c_\emptyset$, $\epsilon_\text{target} = \epsilon_{\theta_\text{dm}}(x_t, c_\emptyset, t)$.
AbConcept \cite{kumari2023ablating} assigns anchor concepts $c^*$ for each erased concept $c_f$, such as $c_f$ is `Van Gogh painting' and $c^*$ is `painting' or $c_f$ is `a photo of Grumpy cat' and $c^*$ is `a photo of cat', $\epsilon_\text{target} = \epsilon_{\theta_\text{dm}}(x_t, c^*, t)$.
In contrast, FMN \cite{zhang2023forget} is an untargeted concept erasure method, which minimizes attention weights corresponding to the forgotten concepts $c_f$.

These advanced approaches focus on unlearning concepts but compromise model performance significantly. Moreover, they have not considered the scenarios of multi-concept erasure and subsequent restoration. 
In this work, we introduce a separable multi-concept erasure framework. It incorporates an untargeted concept-irrelevant erasure mechanism to preserve model performance during concept erasure, and a weight decoupling mechanism to provide flexibility in both the erasure and recovery of concepts.

\section{Proposed Method}
Our proposed separable multi-concept eraser (SepME) aims to flexibly erase or recover multiple concepts while preserving the overall model performance.
An overview of SepME is illustrated in Fig.~\ref{fig1}, which incorporates the generation of concept-irrelevant representations (G-CiRs) and the weight decoupling (WD).

\subsection{G-CiRs}\label{3.1}
To maintain the generative capability of diffusion models (DMs) for regular concepts (or concept-free $c_\emptyset$) during unlearning, G-CiRs prevents the erasure of significant but irrelevant information to forgotten concepts $c_{i,f}\in c_f$, $i\in[1,\text{N}]$, where $\text{N}$ is the number of erased concepts.
Specifically, given original DMs parameterized by $\theta_\text{dm}$, we employ the noise difference $\Delta_\epsilon(c_{i,f},\theta_\text{dm})$ = $\epsilon_{\theta_\text{dm}}(x_t, c_{i,f}, t) - \epsilon_{\theta_\text{dm}}(x_t, c_\emptyset, t)$ to represent the concept $c_{i,f}$.
To successfully erase concepts $c_f$ from DMs, 
the representations of concepts $c_f$ for unlearned and original DMs should be uncorrelated, that is
$\forall_{i\in[1,\text{N}]}\mathcal{L}_\text{cor}(c_{i,f},{\bf \Delta}\theta_\text{dm})$ == 0,
\begin{equation}\label{eq62}
\small
    \mathcal{L}_\text{cor}(c_{i,f},{\bf \Delta}\theta_\text{dm}) = \text{Avg}(\Delta_\epsilon(c_{i,f},\theta_\text{dm})\odot \Delta_\epsilon(c_{i,f}, \theta_\text{dm}+{\bf \Delta}\theta_\text{dm})),
\end{equation}
where ${\bf \Delta}\theta_\text{dm}$ represents the learnable weight increments of unlearned DMs, $\odot$ denotes the element-wise product, and $\text{Avg}(\cdot)$ calculates the average value.
Eq.~(\ref{eq62}) actually computes the relevance between two representations for the concept $c_{i,f}$.
On this basis, we fine-tune ${\bf \Delta}\theta_\text{dm}$ by
\begin{equation}\label{eq61}
\begin{aligned}
    \min_{{\bf \Delta}\theta_\text{dm}}&\mathcal{L}_\text{G-CiRs} = \sum_{i=1}^\text{N}\eta_i\mathcal{L}_\text{cor}(c_{i,f},{\bf \Delta}\theta_\text{dm}) + \lambda\|{\bf \Delta}\theta_\text{dm}\|_p, \\
    &s.t.\; \forall_{i\in[1,\text{N}]}\mathcal{L}_\text{cor}(c_{i,f},{\bf \Delta}\theta_\text{dm}) == 0,
\end{aligned}
\end{equation}
where $\eta_i$ is used to balance the losses of multiple concepts, $\eta_i = \frac{\|\mathcal{L}_\text{cor}(c_{1,f},{\bf \Delta}\theta_\text{dm})\|_2}{\|\mathcal{L}_\text{cor}(c_{i,f},{\bf \Delta}\theta_\text{dm})\|_2}$. $\lambda$ denotes the hyperparameter. $\|{\bf \Delta}\theta_\text{dm}\|_p$ restricts the weight deviation of the unlearned DMs from the original ones.

To satisfy the condition of zero relevance
in Eq.~(\ref{eq61}), we utilize the momentum statistic method 
since the values of $\mathcal{L}_\text{cor}(c_{i,f},{\bf \Delta}\theta_\text{dm})$ computed from various noised samples $x_t$ exhibit significant variations. Specifically, early stopping is activated once $\mathcal{L}_\text{mom}^n \leq \tau$, where $\tau$ denotes the threshold, with a default value of 0.
\begin{equation}\label{eq7}
\begin{aligned}
    \mathcal{L}_\text{mom}^n = \alpha\mathcal{L}_\text{mom}^{n-1} + (1-\alpha)\sum_{i=1}^\text{N}\eta_i\mathcal{L}_\text{cor}(c_{i,f},{\bf \Delta}\theta_\text{dm}),
\end{aligned}
\end{equation}
where $\alpha$ is the hyper-parameter and $n$ is the unlearning step.


\subsection{Weight Decoupling (WD)}
To resolve the subsequent restoration issue of multi-concept erasure,
we decompose   the weights ${\bf \Delta}\theta_\text{dm}$ in Eq.~(\ref{eq62})  into ${\bf \Delta}\theta_{{1\sim \text{N}, \text{dm}}}$ for flexibly manipulating various concepts.
Each independent weight increment ${\bf \Delta}\theta_{{i,\text{dm}}}$ aims to unlearn a specific concept $c_{i,f}$ without compromising the generation performance of models regarding other concepts.
The process of separable erasure can be formulated as:
\begin{equation}\label{eq8}
\begin{aligned}
    \min_{{\bf \Delta}\theta_{{1\sim \text{N}, \text{dm}}}}&\mathcal{L}_\text{SepME} = \sum_{i=1}^\text{N}\eta_i\mathcal{L}_\text{cor}(c_{i,f}, {\bf \Delta}\theta_{{i, \text{dm}}}) + \lambda\|{\bf \Delta}\theta_{{1\sim \text{N}, \text{dm}}}\|_p, \\
    &s.t.\; \forall_{i\in[1,\text{N}]}\, \text{cond}(c_{i,f}, {\bf \Delta}\theta_{{i, \text{dm}}}),
\end{aligned}
\end{equation}
where the conditions 
include:
\begin{equation}\label{eq9}
    \begin{aligned}
    &\mathcal{L}_\text{cor}(c_{i,f},{\bf \Delta}\theta_{{i, \text{dm}}}) == 0,\\
    & \epsilon_{\theta_\text{dm}}(x_t, c_\emptyset, t) == \epsilon_{\theta_\text{dm}+ {\bf \Delta}\theta_{{i,\text{dm}}}}(x_t, c_\emptyset, t),\\
    & \forall_{j\in [1,\text{N}], j\neq i} \epsilon_{\theta_\text{dm}}(x_t, c_{j,f}, t) == \epsilon_{\theta_\text{dm}+ {\bf \Delta}\theta_{{i,\text{dm}}}}(x_t, c_{j,f}, t), \\
    & \forall_{c_{\not\in f}}\epsilon_{\theta_\text{dm}}(x_t, c_{\not\in f}, t) \approx \epsilon_{\theta_\text{dm}+ {\bf \Delta}\theta_{{i,\text{dm}}}}(x_t, c_{\not\in f} , t),
    \end{aligned}    
\end{equation}
where $x_t$ can be an arbitrary image and $c_{\not\in f}$ represents concepts that do not belong to $c_f$.
The first condition aims to erase forgotten concepts, while the rest mitigate the impact of fine-tuning on other concepts.

To meet these conditions, we first need to identify the nonzero positions for ${\bf \Delta}\theta_{{i,\text{dm}}}$.
It is evident from Eq. (\ref{eq9}) that these positions are image-independent. 
Consequently, only the to$\_$k and to$\_$v layers of the cross-attention modules are selected, which are exclusively designed for extracting text embeddings in DMs.
Other positions, such as the to$\_$q layer for extracting image embeddings and the FFN (Feed-Forward Network) for updating fused embeddings, have been fixed.
For $\forall_{i\in[1,\text{N}]}{\bf \Delta}\theta_{{i, \text{dm}}}$, they share the same nonzero positions but have distinct values.

Next, we take ${\bf \Delta}\theta_{\text{to}\_\text{k}}$ as an example to analyze how to determine its value, where $\forall {\bf \Delta}\theta_{\text{to}\_\text{k}}\in {\bf \Delta}\theta_{{i, \text{dm}}}$.
Notably, $\epsilon_{\theta_\text{dm}}(x_t, c_k, t) == \epsilon_{\theta_\text{dm}+ {\bf \Delta}\theta_{\text{to}\_\text{k}}}(x_t, c_k, t)$ means $c_k\otimes {\bf \Delta}\theta_{\text{to}\_\text{k}} $ == 0, where
$c_{k}\in\mathbb{R}^{d_\text{emb}\times d_\text{in}}$, ${\bf \Delta}\theta_{\text{to}\_\text{k}}$ $\in$ $\mathbb{R}^{d_\text{in}\times d_\text{out}}$. $d_\text{emb}$, $d_\text{in}$ and $d_\text{out}$ indicate feature dimensions. $\otimes$ denotes matrix multiplication. 
Thus, Eq. (\ref{eq9}) can be rewritten as
\begin{equation}\label{eq10}
    \begin{aligned}
    &\mathcal{L}_\text{cor}(c_{i,f},{\bf \Delta}\theta_{{i, \text{dm}}}) == 0,\\
    & c_\emptyset \otimes {\bf \Delta}\theta_{\text{to}\_\text{k}} == 0,\\ 
    & \forall_{j\in [1,\text{N}], j\neq i}\, c_{j,f} \otimes {\bf \Delta}\theta_{\text{to}\_\text{k}} == 0,\\    
    &\forall_{c_{\not\in f}}\, c_{\not\in f}\otimes {\bf \Delta}\theta_{\text{to}\_\text{k}}\approx 0^{d_\text{emb}\times d_\text{out}}\Rightarrow {\bf \Delta}\theta_{\text{to}\_\text{k}}\approx 0^{d_\text{in}\times d_\text{out}}.\\
    \end{aligned}    
\end{equation}

$\blacklozenge$
To make ${\bf \Delta}\theta_{\text{to}\_\text{k}}$  satisfy the second and third conditions in Eq.~(\ref{eq10}),
we first compute the particular solutions $S_p$ to a system of linear equations $A\otimes {S_p} = 0$,
\begin{equation}\label{eq11}
    A = [c_\emptyset^\top; c_{1,f}^\top; \cdots; c_{i-1,f}^\top; c_{i+1,f}^\top; \cdots; c_{\text{N},f}^\top]^\top,
\end{equation}
where $A$ is a constant matrix for specified concepts $c_{f}$, ${A}\in\mathbb{R}^{(\text{N}\cdot d_\text{emb})\times d_\text{in}}$.
$\top$ means matrix transpose.
$S_p\in\mathbb{R}^{d_\text{in}\times d_\text{in} - r}$, 
where $r$ is the rank of $A$, with $r\leq \text{N}\cdot d_\text{emb}$. 
$d_\text{in} - r$ quantifies the number of solutions within $S_p$. 
$d_\text{in} \gg d_\text{emb}$ in DMs.
Notably, to remove the original biases in solutions $S_p$, we normalize each element of $S_p$ to a unit vector.

Then, each column of ${\bf \Delta}\theta_{\text{to}\_\text{k}}$ can be formulated as a linear combination of these solutions $S_p$,
\begin{equation}\label{eq13}
     {\bf \Delta}\theta_{\text{to}\_\text{k}} = (w \otimes S_p^\top)^\top,
\end{equation}
where $w$ is an optimizable variable and represents the linear combination weights, $w\in\mathbb{R}^{d_\text{out}\times(d_\text{in}-r)}$.

$\blacklozenge$ To further make ${\bf \Delta}\theta_{\text{to}\_\text{k}}$  satisfy the fourth condition in Eq.~(\ref{eq10}), we introduce a scaling factor $\beta$ as follows,
\begin{equation}\label{eq14}
     {\bf \Delta}\theta_{\text{to}\_\text{k}} = (w \otimes (\beta S_p^\top))^\top.
\end{equation}
Meanwhile, $w$ is initialized to a zero matrix.
Additionally, we replace $\|{\bf \Delta}\theta_{{1\sim \text{N}, \text{dm}}}\|_p$ in Eq.~(\ref{eq8}) with the $\|\mathcal{W}\|_p$.
Here, $\mathcal{W}$ represents the set of optimizable variables $w$ defined for both to$\_$k and to$\_$v layers.

Overall, the objective of SepME is simplified as:
\begin{equation}\label{eq15}
\begin{aligned}
    \min_\mathcal{W}&\mathcal{L}_\text{SepME} = \sum_{i=1}^\text{N}\eta_i\mathcal{L}_\text{cor}(c_{i,f},{\bf \Delta}\theta_{{i, \text{dm}}}) + \lambda\|\mathcal{W}\|_p, \\
    &s.t.\; \forall_{i\in[1,\text{N}]}\mathcal{L}_\text{cor}(c_{i,f},{\bf \Delta}\theta_{{i, \text{dm}}}) == 0.
\end{aligned}
\end{equation}

{\bf Evaluation for SepME.} We combine various ${\bf \Delta}\theta_{{i, \text{dm}}}$ to erase the corresponding concepts. For instance, DMs with $\theta_{\text{dm}} + \sum_{i\in \{j,k\}}{\bf \Delta}\theta_{{i, \text{dm}}}$ eliminate the concepts $c_{j,f}$ and $c_{k,f}$.

\section{Experiments}
\subsection{Experimental Settings}\label{sec4.1}
{\bf Implementation Details.} 
We follow prior works \cite{gandikota2023erasing,kim2023towards} to fine-tune Stable Diffusion \cite{rombach2022high}.
The optimization process utilizes the Adam optimizer for a maximum of 1000 iterations and our early stopping strategy.
The batch size is set equal to the number of erased concepts. 
When exclusively evaluating the G-CiRs module, the learning rate is set to 1e-6, and we opt to fine-tune the cross-attention modules. 
For assessing the SepME, the learning rate is adjusted to 1e-2, and optimization is conducted on the to$\_$k and to$\_$v layers of the cross-attention modules. 
The default values for hyperparameters $\tau$, $\alpha$ in Eq.~(\ref{eq7}), $\beta$ in Eq.~(\ref{eq14}), and $\lambda$ in Eq.~(\ref{eq15}) are set to 0, 0.9, 1e-4, and 3e-5, respectively.
The threshold $\tau$ controls the moment of early stopping, and ablation studies for $\tau$ are provided in the appendix.
All experiments are executed on 2 RTX 3090 GPUs.

\begin{table*}[t]
    \begin{center}
    \caption{Quantitative results of the single concept erasure. `VG', `PC' and `CE' are artists of `Van Gogh', `Picasso' and `Cezanne', respectively. $\bar{\uparrow}$: Since the unlearning aims to erase the concept style, an intermediate value may indicate better performance. 
    $i$ in FMN$_i$ represents the iteration step. Text in \textcolor{red!100}{red} and \textcolor{blue!100}{blue} denotes the best and second-best results, respectively.
    }
    \label{tab1}
    \begin{tabular}{c|c|cccccc}
    \hline
    \multicolumn{2}{c|}{ACC/LPIPS}&\multicolumn{6}{c}{Unlearning methods}\\
    \hline
Erased &Evaluated &ORI&FMN$_{10}$&FMN$_{20}$&ESD&AbConcept &Ours-G-CiRs  \\
         \midrule
         \multirow{3}{*}{VG}
         &VG$^*$($\downarrow/\bar{\uparrow}$)
         &1.000/0.000&0.544/0.433&0.500/0.517&0.320/0.472&\textcolor{red!100}{0.000}/0.469&\textcolor{blue!100}{0.228}/0.363\\
         &PC$^\dagger$($\uparrow/\downarrow$)
         &1.000/0.000&\textcolor{red!100}{1.000}/\textcolor{blue!100}{0.186}&\textcolor{red!100}{1.000}/0.256&\textcolor{red!100}{1.000}/0.301&0.900/0.190&\textcolor{red!100}{1.000}/\textcolor{red!100}{0.175}\\
         &CE$^\dagger$($\uparrow/\downarrow$)
         &1.000/0.000&\textcolor{red!100}{1.000}/0.178&0.964/0.257&\textcolor{red!100}{1.000}/\textcolor{blue!100}{0.164}&0.820/0.217&\textcolor{red!100}{1.000}/\textcolor{red!100}{0.154}\\
         \hline
         \multirow{3}{*}{PC}
         &VG$^\dagger$($\uparrow/\downarrow$)
         &1.000/0.000&\textcolor{red!100}{1.000}/\textcolor{red!100}{0.205}&\textcolor{blue!100}{0.952}/0.265&0.908/0.209&0.684/0.227&0.908/\textcolor{red!100}{0.180}\\
         &PC$^*$($\downarrow/\bar{\uparrow}$)
         &1.000/0.000&{1.000}/0.231&0.952/0.339&0.400/0.424&\textcolor{red!100}{0.000}/0.397&\textcolor{blue!100}{0.052}/0.358\\
         &CE$^\dagger$($\uparrow/\downarrow$)
         &1.000/0.000&\textcolor{red!100}{1.000}/\textcolor{red!100}{0.149}&\textcolor{red!100}{1.000}/\textcolor{blue!100}{0.175}&\textcolor{red!100}{1.000}/0.211&0.752/0.194&\textcolor{red!100}{1.000}/0.189\\
         \hline
         \multirow{3}{*}{CE}
         &VG$^\dagger$($\uparrow/\downarrow$)
         &1.000/0.000&\textcolor{red!100}{0.952}/\textcolor{red!100}{0.201}&0.772/0.305&\textcolor{blue!100}{0.908}/\textcolor{blue!100}{0.238}&0.728/0.257&\textcolor{blue!100}{0.908}/0.276\\
         &PC$^\dagger$($\uparrow/\downarrow$)
         &1.000/0.000&\textcolor{red!100}{1.000}/\textcolor{red!100}{0.184}&\textcolor{red!100}{1.000}/0.207&\textcolor{red!100}{1.000}/0.314&0.852/0.251&\textcolor{red!100}{1.000}/0.204\\
         &CE$^*$($\downarrow/\bar{\uparrow}$)
         &1.000/0.000&{1.000}/0.189&0.820/0.351&0.428/0.372&\textcolor{blue!100}{0.036}/0.360&\textcolor{red!100}{0.000}/0.363\\
         \midrule
         \multicolumn{2}{c|}{$\sum \cdot^* -\sum \cdot^\dagger$($\downarrow/\bar{\uparrow}$)}&-&-3.41/-0.25&-3.42/-0.26&-4.67/{-0.17}&\textcolor{blue!100}{-4.70}/-0.11&\textcolor{red!100}{\bf -5.54}/{-0.09}\\
         \hline
         \multicolumn{2}{c|}{$\|{\bf \Delta}\theta_\text{dm}\|_p\downarrow$}&ORI&FMN$_{10}$&FMN$_{20}$&ESD&AbConcept&Ours-G-CiRs\\\hline
         \multicolumn{2}{c|}{VG}&0.000&\textcolor{blue!100}{44.18}&86.05&120.8&158.3&\textcolor{red!100}{\bf 43.17}\\
         \multicolumn{2}{c|}{PC}&0.000&\textcolor{blue!100}{45.41}&93.63&126.9&146.5&\textcolor{red!100}{\bf 36.90}\\
         \multicolumn{2}{c|}{CE}&0.000&\textcolor{blue!100}{45.31}&92.32&128.5&156.9&\textcolor{red!100}{\bf 36.55}\\
         \hline
    \end{tabular}
    \end{center}
    \vspace{-2mm}
\end{table*}

\begin{figure}[t]
    \begin{center}
        \includegraphics[width=1\linewidth]{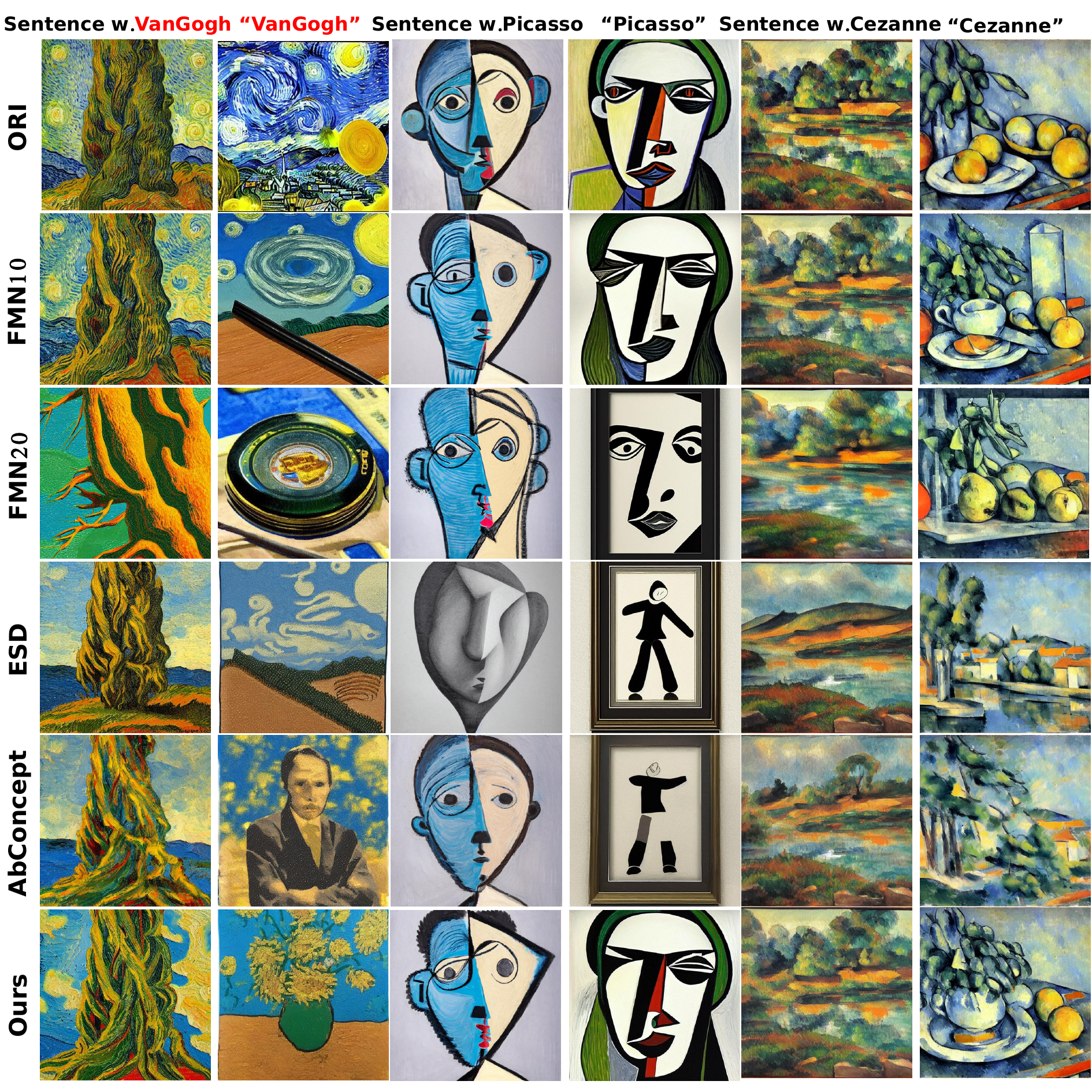}
    \end{center}
    \vspace{-2mm}
    \caption{    
    Qualitative comparison among various unlearning techniques for DMs with `Van Gogh' as the erased concept.
    }
    \label{fig2}
    \vspace{-4mm}
\end{figure}

\begin{figure}[t]
    \begin{center}
        \includegraphics[width=1\linewidth]{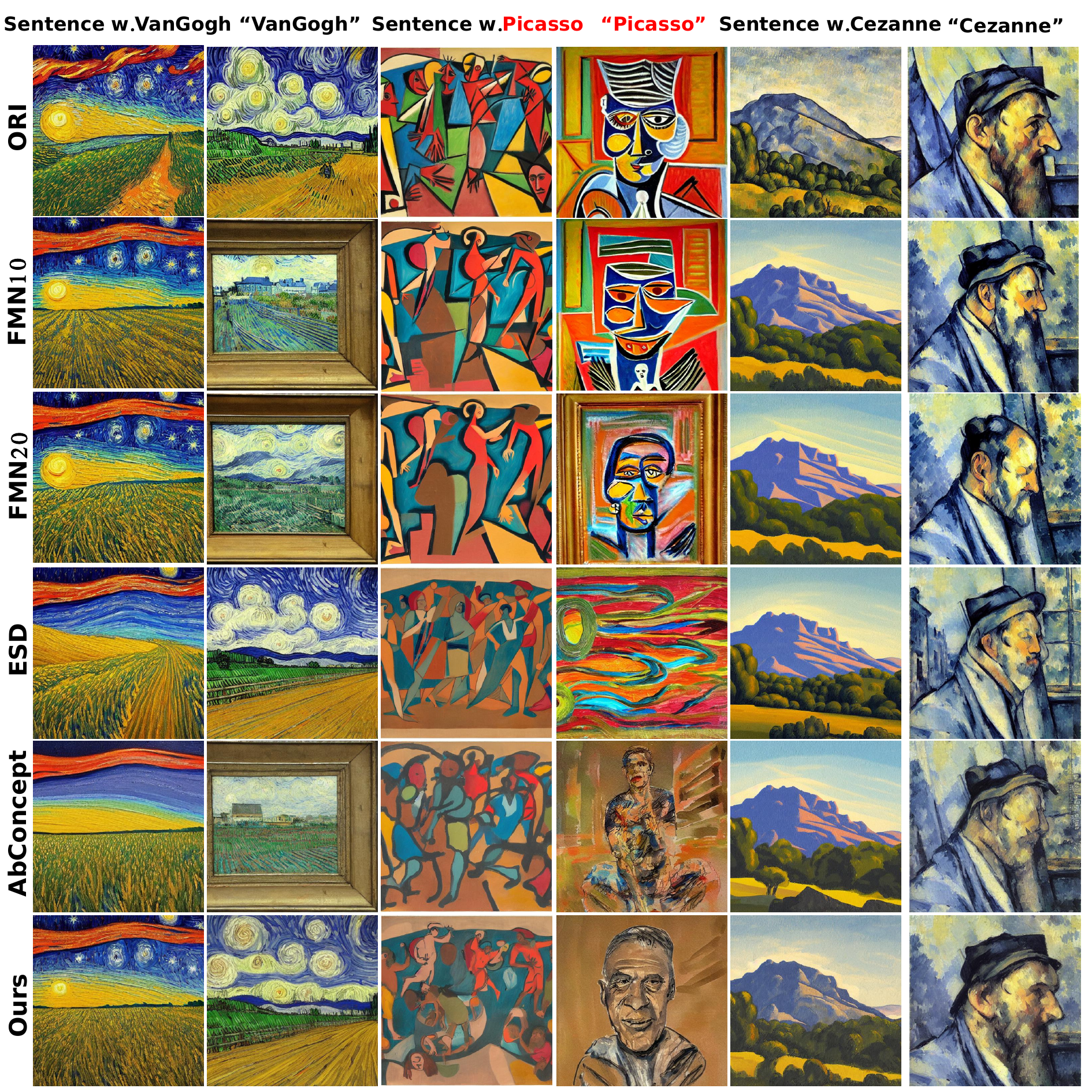}
    \end{center}
    \vspace{-2mm}
    \caption{    
    Qualitative comparison among various unlearning techniques for DMs with `Picasso' as the erased concept.
    }
    \label{fig3}
    \vspace{-4mm}
\end{figure}

{\bf Evaluation metrics.}
The evaluation metrics include modifications to model parameters $\|{\bf \Delta}\theta_\text{dm}\|_p = \frac{\|{\bf \Delta}\theta_\text{dm}\|_1}{\text{M}}$, perceptual distance measured by Perceptual Image Patch Similarity (LPIPS), and classification accuracy (ACC). Here, $\text{M}$ denotes the number of layers.
LPIPS quantifies the similarity between the original image and the image generated by unlearned DMs, calculated based on AlexNet \cite{krizhevsky2012imagenet} with settings from the source code \footnote{\url{https://github.com/richzhang/PerceptualSimilarity}}.

\begin{table*}[t]
    \tabcolsep = 0.1cm
    \begin{center}
    \caption{Quantitative results of the multi-concept erasure. `VG', `PC' and `CE' are `Van Gogh', `Picasso' and `Cezanne', respectively.
    $\bar{\uparrow}$: Since the unlearning aims to erase the concept style, an intermediate value may indicate better performance.
    $i$ in FMN$_i$ represents the iteration step. Text in \textcolor{red!100}{red} and \textcolor{blue!100}{blue} denotes the best and second-best results, respectively.}
    \label{tab2}
    \begin{tabular}{c|c|ccccccc}
    \hline
    \multicolumn{2}{c|}{ACC/LPIPS}&\multicolumn{7}{c}{Unlearning methods}\\
    \hline
Erased &Evaluated&ORI&FMN$_{20}$&FMN$_{30}$&FMN$_{50}$&ESD&AbConcept&Ours-G-CiRs  \\
         \midrule
         \multirow{4}{*}{VG+PC+CE}
         &VG$^*$($\downarrow/\bar{\uparrow}$)&1./0.&0.544/0.487&0.500/0.511&\textcolor{blue!100}{0.136}/0.555&0.364/0.467&\textcolor{red!100}{0.092}/0.421&0.224/0.426\\
         &PC$^*$($\downarrow/\bar{\uparrow}$)&1./0.&0.952/0.299&0.552/0.299&\textcolor{red!100}{0.000}/0.436&0.252/0.442&\textcolor{blue!100}{0.132}/0.329&\textcolor{red!100}{0.000}/0.359\\
         &CE$^*$($\downarrow/\bar{\uparrow}$)&1./0.&\textcolor{blue!100}{0.180}/0.359&\textcolor{red!100}{0.000}/0.424&\textcolor{red!100}{0.000}/0.505&0.356/0.343&0.500/0.286&\textcolor{red!100}{0.000}/0.419\\
         &Others$^\dagger$($\uparrow/\downarrow$)&1./0.&0.955/0.228&0.878/0.269&0.693/0.358&0.897/0.252&\textcolor{red!100}{0.977}/\textcolor{red!100}{0.198}&\textcolor{blue!100}{0.958}/\textcolor{blue!100}{0.223}\\
         \midrule
         \multicolumn{2}{c|}{$\sum \cdot^* -\sum \cdot^\dagger$($\downarrow/\bar{\uparrow}$)}&-&0.721/0.917&-0.17/0.965&\textcolor{blue!100}{\bf -0.56}/1.138&-0.07/1.000&-0.26/0.838&\textcolor{red!100}{\bf -0.73}/0.981\\
         \hline
         \multicolumn{2}{c|}{$\|{\bf \Delta}\theta_\text{dm}\|_p\downarrow$}&ORI&FMN$_{20}$&FMN$_{30}$&FMN$_{50}$&ESD&AbConcept&Ours-G-CiRs\\\hline
         \multicolumn{2}{c|}{VG+PC+CE}&0.0&\textcolor{blue!100}{92.58}&150.3&243.5&128.4&153.8&\textcolor{red!100}{\bf 58.22}\\
         \hline
    \end{tabular}
    \end{center}
    \vspace{-2mm}
\end{table*}
\begin{figure}[t]
    \begin{center}
        \includegraphics[width=1\linewidth]{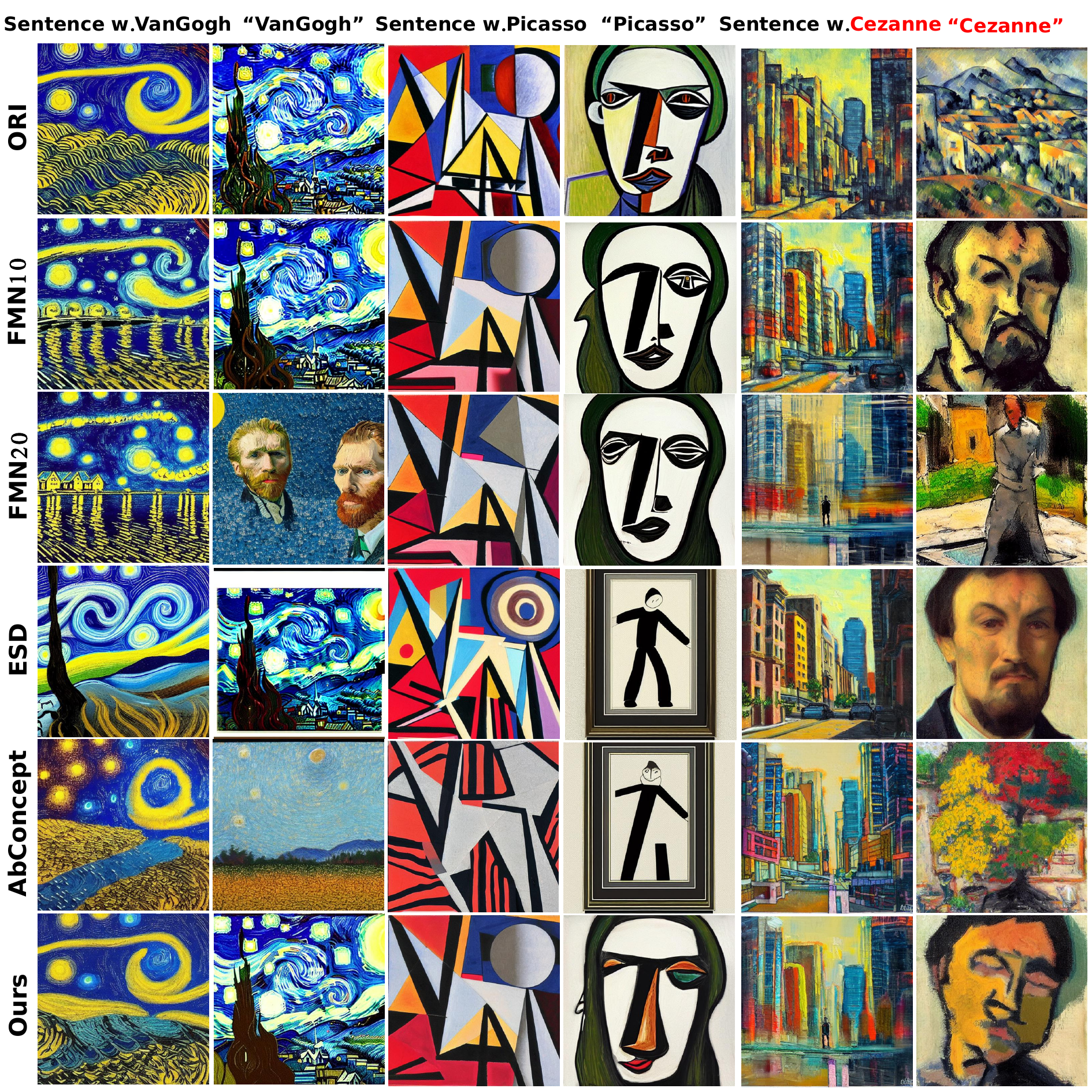}
    \end{center}
    \vspace{-2mm}
    \caption{    
    Qualitative comparison among various unlearning techniques for DMs with `Cezanne' as the erased concept.
    }
    \label{fig4}
    \vspace{-2mm}
\end{figure}

\begin{figure}[t]
    \begin{center}
        \includegraphics[width=1\linewidth]{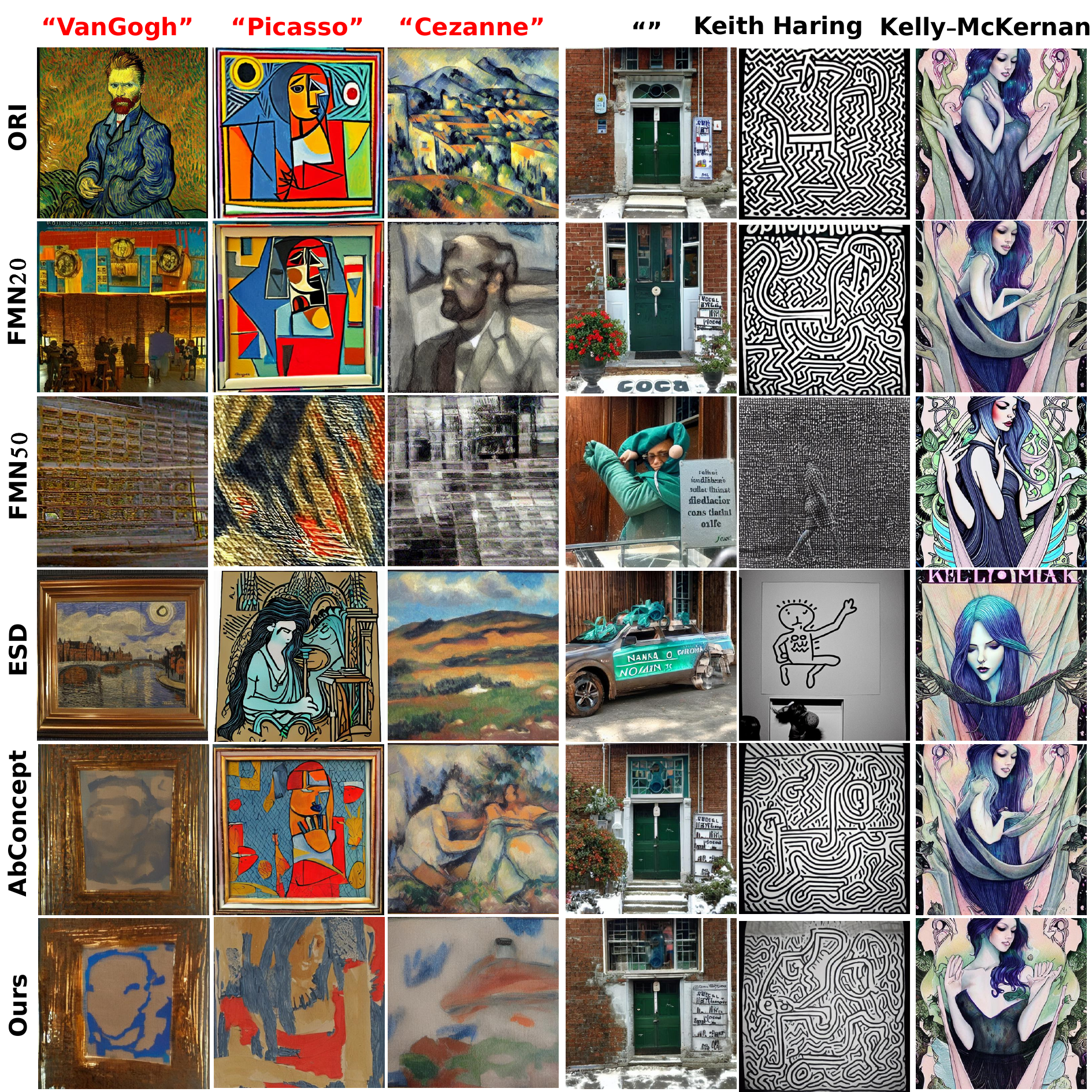}
    \end{center}
    \vspace{-2mm}
    \caption{    
    Qualitative comparison among various unlearning techniques under multi-concept erasure. `Ours' indicates G-CiRs.
    }
    \label{fig5}
    \vspace{-2mm}
\end{figure}

We calculate ACC using various pre-trained classification models. 
For the style classification model, we consider a blank concept and nine artist styles: `Van Gogh', `Picasso', `Cezanne', `Jackson Pollock', `Caravaggio', `Keith Haring', `Kelly McKernan', `Tyler Edlin', and `Kilian Eng'.
In each category, we generate 1000 images using the original DMs with artist names (or `') as prompts. 70\% of the data is allocated for training purposes, while the remaining 30\% is reserved for testing. Only the fully connected (FC) layer of the pre-trained ResNet18 model \cite{he2016deep} is optimized with 20 epochs.
The cyclical learning rate \cite{smith2017cyclical} is employed with the maximum learning rate of 0.01.
As for the object classification network, we directly utilize the pre-trained ResNet50.

Advanced unlearning methods, including FMN \cite{zhang2023forget}, ESD \cite{gandikota2023erasing} and AbConcept \cite{kim2023towards}, are used as {\bf baselines}.

\subsection{Style Removal}
To evaluate the performance of our methods in eliminating styles, we focus on three prominent artists: `Van Gogh', `Picasso', and `Cezanne'. 
During fine-tuning, we employ the images for obtaining the style classification model as inference images $x_0$ to yield $x_t$, {\em i.e.,} $x_t = \sqrt{\Bar{\alpha_t}}x_0 + \sqrt{1-\Bar{\alpha_t}}\epsilon$.
During evaluation, we generate 250 images for each concept, {\em i.e.,} 50 seeds for each concept, and 5 images per seed.
As our SepME relies on the proposed G-CiRs module, we first indicate the performance of G-CiRs and then validate the efficacy of SepME.

\subsubsection{Evaluation for G-CiRs} 

For the \textit{single concept erasure}, one style is chosen as the forgotten concept, while others serve as evaluation concepts.
The quantitative results are presented in Tab.~\ref{tab1}.
\textcolor{blue!100}{On one hand, our G-CiRs achieves optimal performance in terms of ACC and LPIPS metrics across three artistic styles. On the other hand, the proposed G-CiRs induces fewer modifications to model weights.} 
Furthermore, qualitative comparisons in Figs.~\ref{fig2}$\sim$\ref{fig4} also demonstrate the efficacy of our method in erasing artistic style and preserving generative performance across other concepts.
These visual samples are produced with artist names or sentences \footnote{\url{https://github.com/rohitgandikota/erasing/blob/main/data/art_prompts.csv}} 
containing artist names as prompts for various unlearned DMs.

\begin{table*}[htpb]
    \tabcolsep = 0.05cm
    \begin{center}
    \caption{Quantitative results of SepME when simultaneously fine-tuning ${\bf \Delta}\theta_{{1\sim 3, \text{dm}}}$. `VG', `CE' and `PC' are artists of `Van Gogh', `Cezanne' and `Picasso', respectively. ${\bf \Delta}\theta_{{1, \text{dm}}}$, ${\bf \Delta}\theta_{{2, \text{dm}}}$, ${\bf \Delta}\theta_{{3, \text{dm}}}$ are optimizable weights for erasing `VG', `CE' and `PC', respectively.
    $i$ in FMN$_i$ represents the iteration step. 
    For each combination, we employ AbConcept and G-CiRs as baselines to re-finetune all layers of the cross-attention module in DMs to eliminate the corresponding concepts. Text in \textcolor{red!100}{red} indicates the best result.
    }
    \label{tab3}
    \begin{tabular}{c|cccccccc}
    \hline
    &\multicolumn{7}{c}{SepME(ACC/LPIPS)}\\
$\theta_\text{dm}$&+0&$+{\bf \Delta}\theta_{{1, \text{dm}}}$&$+{\bf \Delta}\theta_{{2, \text{dm}}}$&$+{\bf \Delta}\theta_{{3, \text{dm}}}$&$+\sum_{i=1}^2{\bf \Delta}\theta_{{i, \text{dm}}}$&$+\sum_{i\in\{1,3\}}{\bf \Delta}\theta_{{i, \text{dm}}}$  &$+\sum_{i=2}^3{\bf \Delta}\theta_{{i, \text{dm}}}$&$+\sum_{i=1}^3{\bf \Delta}\theta_{{i, \text{dm}}}$\\
         \hline
         VG&1./0.&0.320/0.371&\textcolor{red!100}{0.956}/0.182&\textcolor{red!100}{0.956}/0.182&0.364/0.372&0.272/0.371&\textcolor{red!100}{0.956}/0.182&0.364/0.372\\
         CE&1./0.&\textcolor{red!100}{1.000}/0.144&0.180/0.303&\textcolor{red!100}{1.000}/0.144&0.180/0.303&\textcolor{red!100}{1.000}/0.144&0.180/0.303&0.180/0.303\\
         PC&1./0.&\textcolor{red!100}{1.000}/0.185&\textcolor{red!100}{1.000}/0.185&\textcolor{red!100}{0.000}/0.440&\textcolor{red!100}{1.000}/0.185&\textcolor{red!100}{0.000}/0.440&\textcolor{red!100}{0.000}/0.440&\textcolor{red!100}{0.000}/0.440\\
        \hline
        &\multicolumn{7}{c}{AbConcept(ACC/LPIPS)}\\
        $c_f$&-&VG&CE&PC&VG+CE&VG+PC&CE+PC&VG+CE+PC\\
        VG&1./0.&\textcolor{red!100}{0.000}/0.469&0.728/0.257&0.684/0.227&\textcolor{red!100}{0.136}/0.402&\textcolor{red!100}{0.044}/0.430&0.728/0.257&\textcolor{red!100}{0.000}/0.425\\
        CE&1./0.&0.820/0.217&0.036/0.360&0.752/0.194&0.572/0.253&0.820/0.194&0.252/0.298&0.464/0.269\\
        PC&1./0.&0.900/0.190&0.852/0.251&\textcolor{red!100}{0.000}/0.397&0.952/0.172&0.352/0.284&0.728/0.257&0.152/0.301\\
        \hline
        &\multicolumn{7}{c}{G-CiRs(ACC/LPIPS)}\\
        $c_f$&-&VG&CE&PC&VG+CE&VG+PC&CE+PC&VG+CE+PC\\
        VG&1./0.&0.228/0.363&0.908/0.180&0.908/0.276& 0.184/0.384&0.184/0.382&0.700/0.259&0.224/0.426\\
        CE&1./0.&\textcolor{red!100}{1.000}/0.154&\textcolor{red!100}{0.000}/0.363&\textcolor{red!100}{1.000}/0.189&\textcolor{red!100}{0.108}/0.354&0.780/0.173&\textcolor{red!100}{0.036}/0.427&\textcolor{red!100}{0.000}/0.359\\
        PC&1./0.&\textcolor{red!100}{1.000}/0.175&\textcolor{red!100}{1.000}/0.204&0.052/0.358&1.000/0.210&0.152/0.295&\textcolor{red!100}{0.000}/0.449&\textcolor{red!100}{0.000}/0.419\\
         \hline
    \end{tabular}
    \end{center}
    \vspace{-6mm}
\end{table*}

Likewise, we compare our G-CiRs with previous works under the \textit{simultaneous erasure of multiple concepts}.
Specifically, we consider the styles `Van Gogh,' `Picasso,' and `Cezanne' as forgotten concepts and include additional styles in Sec.~\ref{sec4.1} for evaluation purposes.
The experimental results in Tab.~\ref{tab2} \textcolor{blue!100}{demonstrate that our method achieves optimal performance when simultaneously erasing multiple concepts.}
Furthermore, we provide visual examples in Fig.~\ref{fig5}.
As observed, even when using forgotten concepts as prompts, our G-CiRs does not generate images containing erased styles.
Notably, these generated images contain few discernible objects, which is expected given that the proposed G-CiRs is an untargeted unlearning technique.

\subsubsection{ Evaluation for SepME}
The preceding experiments assessed the unlearning performance of various approaches. 
In the following, we investigate the concept restoration issue overlooked by these methods under two practical scenarios: unlearning multiple concepts simultaneously or iteratively unlearning multiple concepts. {\bf 1)} The former knows all forgotten concepts for each concept erasure.
{\bf 2)} The latter only has knowledge of previously forgotten concepts at each erasure step.

After unlearning, various weights are randomly combined to erase corresponding concepts, such as $\theta_\text{dm} + {\bf \Delta}\theta_{1,\text{dm}}$, $\theta_\text{dm} + {\bf \Delta}\theta_{2,\text{dm}}$, and $\theta_\text{dm} + {\bf \Delta}\theta_{3,\text{dm}}$ for erasing `Van Gogh,' `Cezanne,' and `Picasso,' respectively. The combination $\theta_\text{dm} + \sum_{i=2}^3{\bf \Delta}\theta_{i,\text{dm}}$ erases the concepts `Cezanne' and `Picasso'. As only the to$\_$k and to$\_$v layers of the cross-attention modules are chosen as nonzero positions for SepME, achieving 
$\mathcal{L}_\text{cor}(\text{`Cezanne'},{\bf \Delta}\theta_{2,\text{dm}}) == 0$ is challenging. Therefore, the threshold $\tau$ for `Cezanne' is adjusted to 1.5e-4.


\begin{table*}[t]
    \tabcolsep = 0.1cm
    \begin{center}
    \caption{Quantitative results of SepME when separately fine-tuning ${\bf \Delta}\theta_{{1\sim 3, \text{dm}}}$. The concepts $c_\text{1,f}$, $c_\text{2,f}$, and $c_\text{3,f}$ correspond to `Van Gogh', `Cezanne', and `Picasso', respectively. ${\bf \Delta}\theta_{{1, \text{dm}}}$, ${\bf \Delta}\theta_{{2, \text{dm}}}$, ${\bf \Delta}\theta_{{3, \text{dm}}}$ are optimizable weights for erasing `VG', `CE' and `PC', respectively. In SepME$_1$, ${\bf \Delta}\theta_{{1\sim 3, \text{dm}}}$ are separately optimized when all forgotten concepts are known. SepME$_2$ follows the mode of iterative concept erasure, {\em i.e.,} the $t$-th erasure step only possesses knowledge ($Kn$) of the previously forgotten concepts.
    }
    \label{tab4}
    \begin{tabular}{c|cc ccc cc}
    \hline
    &\multicolumn{7}{c}{SepME$_1$--({ACC/LPIPS})--$Kn = [c_\text{1,f}; c_\text{2,f}; c_\text{3,f}]$}\\
    $\theta_{{\text{dm}}}$&+${\bf \Delta}\theta_{{1, \text{dm}}}$&+${\bf \Delta}\theta_{{2, \text{dm}}}$&+${\bf \Delta}\theta_{{3, \text{dm}}}$&+$\sum_{i=1}^2{\bf \Delta}\theta_{{i, \text{dm}}}$&+$\sum_{i\in\{1,3\}}{\bf \Delta}\theta_{{i, \text{dm}}}$&+$\sum_{i=2}^3{\bf \Delta}\theta_{{i, \text{dm}}}$&+$\sum_{i=1}^3{\bf \Delta}\theta_{{i, \text{dm}}}$\\
    VG&\textcolor{gray!80}{0.356/0.364}&\textcolor{gray!80}{1.000/0.182}&\textcolor{gray!80}{0.908/0.182}&0.308/0.365&0.308/0.365&\underline{0.956}/0.181&\textcolor{gray!80}{0.308/0.364}\\
    CE&\textcolor{gray!80}{1.000/0.144}&\textcolor{gray!80}{0.320/0.304}&\textcolor{gray!80}{1.000/0.144}&0.320/0.304&\underline{1.000}/0.144&0.288/0.304&\textcolor{gray!80}{0.320/0.304}\\
    PC&\textcolor{gray!80}{1.000/0.185}&\textcolor{gray!80}{1.000/0.185}&\textcolor{gray!80}{0.000/0.460}&\underline{1.000}/0.185&0.000/0.460&0.000/0.460&\textcolor{gray!80}{0.000/0.460}\\
    \hline
    &\multicolumn{7}{c}{SepME$_2$--({ACC/LPIPS})}\\
    $Kn$&$[c_\text{1,f}]$&$[c_\text{1,f};c_\text{2,f}]$&$[c_\text{1,f};c_\text{2,f};c_\text{3,f}]$&-&-&-&-\\
    VG&\textcolor{gray!80}{0.228/0.363}&\textcolor{gray!80}{0.956/0.182}&\textcolor{gray!80}{0.956/0.182}&0.228/0.363&0.228/0.363&\underline{0.956}/0.182&\textcolor{gray!80}{0.228/0.363}\\
    CE&\textcolor{gray!80}{1.000/0.182}&\textcolor{gray!80}{0.000/0.404}&\textcolor{gray!80}{1.000/0.172}&0.000/0.376&\underline{1.000}/0.177&0.000/0.411&\textcolor{gray!80}{0.052/0.394}\\
    PC&\textcolor{gray!80}{0.964/0.150}&\textcolor{gray!80}{1.000/0.144}&\textcolor{gray!80}{0.288/0.277}&\underline{0.964}/0.150&0.108/0.292&0.288/0.277&\textcolor{gray!80}{0.108/0.292}\\
    \hline
    &\multicolumn{7}{c}{Abconcept--({ACC/LPIPS})}\\
    VG&\textcolor{gray!80}{0.000/0.469}&\textcolor{gray!80}{0.728/0.257}&\textcolor{gray!80}{0.684/0.227}&0.000/0.472&0.000/0.469&\underline{0.636}/0.277&\textcolor{gray!80}{0.000/0.487}\\
    CE&\textcolor{gray!80}{0.820/0.217}&\textcolor{gray!80}{0.036/0.360}&\textcolor{gray!80}{0.752/0.194}&0.288/0.299&\underline{0.680}/0.234&0.252/0.288&\textcolor{gray!80}{0.144/0.349}\\
    PC&\textcolor{gray!80}{0.900/0.190}&\textcolor{gray!80}{0.852/0.251}&\textcolor{gray!80}{0.000/0.397}&\underline{0.600}/0.277&0.152/0.304&0.200/0.334&\textcolor{gray!80}{0.052/0.364}\\
    \hline
    &\multicolumn{7}{c}{G-CiRs--({ACC/LPIPS})}\\
    VG&\textcolor{gray!80}{0.228/0.363}&\textcolor{gray!80}{0.908/0.180}&\textcolor{gray!80}{0.908/0.276}&0.184/0.386&0.092/0.382&\underline{0.544}/0.276&\textcolor{gray!80}{0.184/0.410}\\
    CE&\textcolor{gray!80}{1.000/0.154}&\textcolor{gray!80}{0.000/0.363}&\textcolor{gray!80}{1.000/0.189}&0.108/0.261&\underline{0.716}/0.178&0.216/0.313&\textcolor{gray!80}{0.036/0.331}\\
    PC&\textcolor{gray!80}{1.000/0.175}&\textcolor{gray!80}{1.000/0.204}&\textcolor{gray!80}{0.052/0.358}&\underline{1.000}/0.220&0.452/0.247&0.200/0.280&\textcolor{gray!80}{0.200/0.303}\\
    \hline
    \end{tabular}
    \end{center}
    \vspace{-4mm}
\end{table*}

\begin{figure}[t]
    \begin{center}
        \includegraphics[width=1\linewidth]{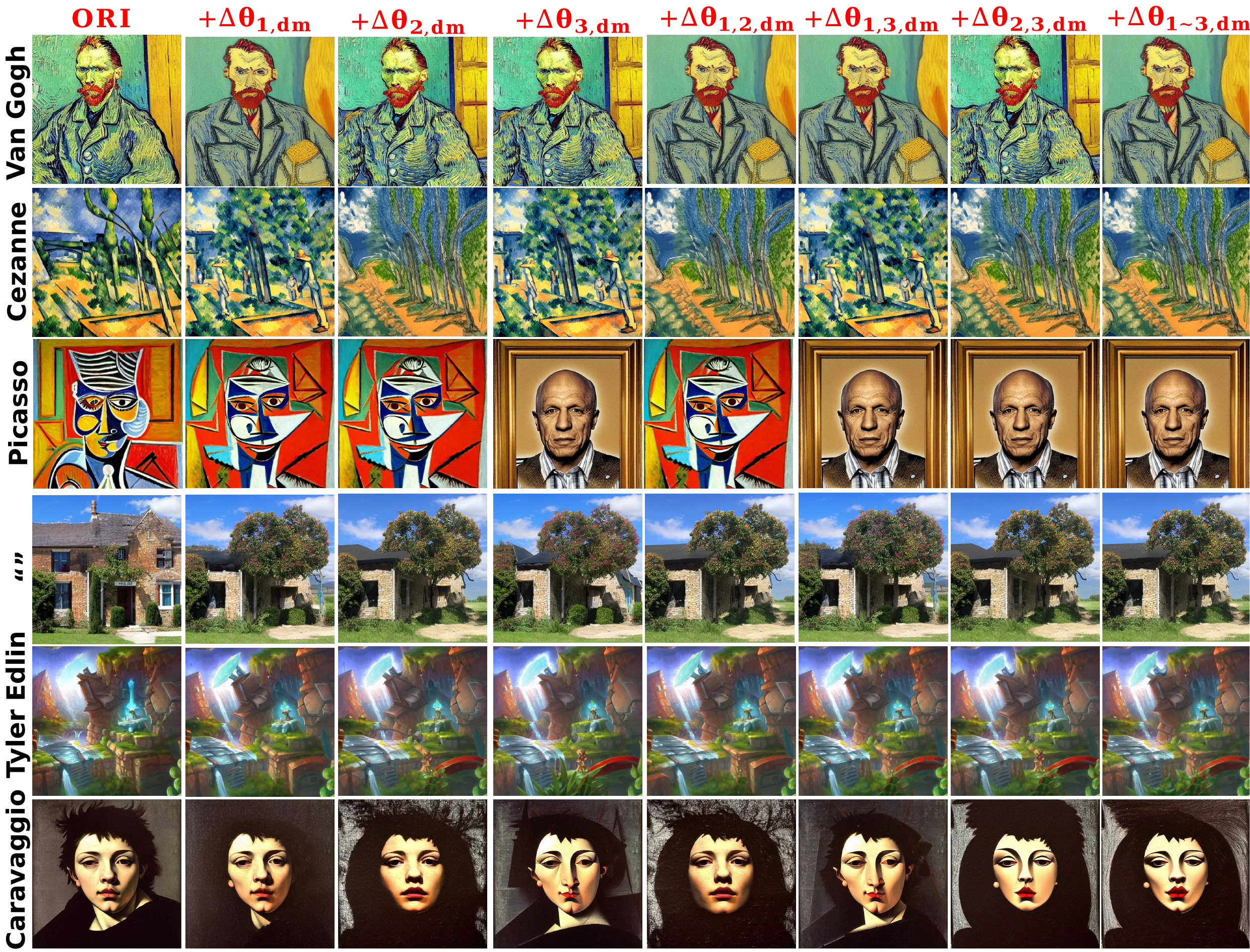}
    \end{center}
    \vspace{-2mm}
    \caption{    
    Visual examples of the proposed SepME.
    }
    \label{fig7}
    \vspace{-2mm}
\end{figure}

\begin{figure}[t]
    \begin{center}
        \includegraphics[width=1\linewidth]{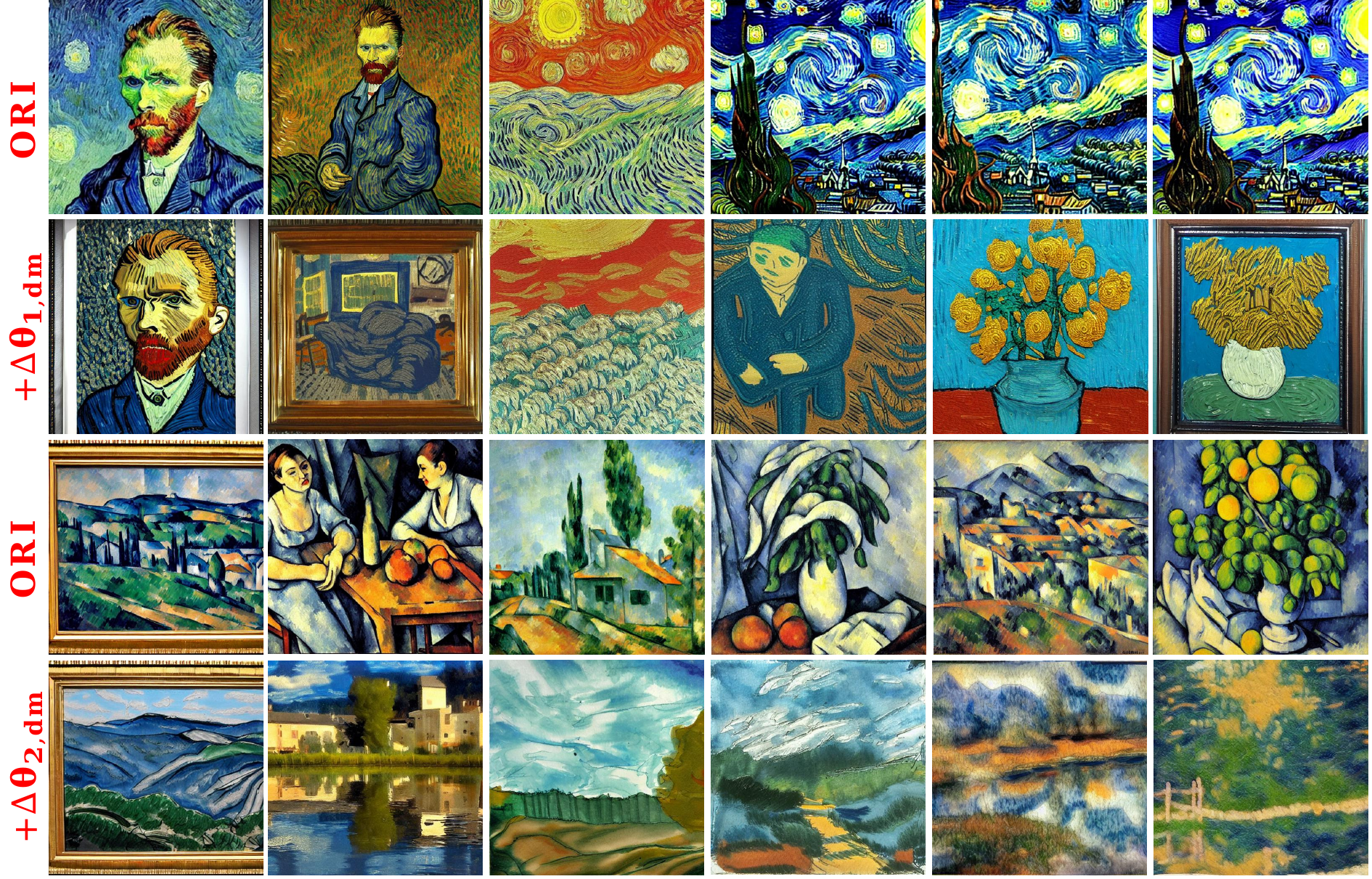}
    \end{center}
    \vspace{-2mm}
    \caption{    
    Failure cases of the proposed SepME.
    }
    \label{fig8}
    \vspace{-2mm}
\end{figure}

{\bf Simultaneous erasure of multiple concepts.}
We first optimize ${\bf \Delta}\theta_{1\sim 3,\text{dm}}$ simultaneously. The quantitative and qualitative results are presented in Tab.~\ref{tab3} and Fig.~\ref{fig7}, respectively. 
For each combination, we utilize AbConcept and G-CiRs as baselines to re-finetune all layers of the cross-attention module in DMs to eliminate the corresponding concepts. \textcolor{blue!100}{It can be observed that SepME can effectively and flexibly erase various concepts, achieving comparable performance to separately fine-tuned methods, such as the results of ${\bf \Delta}\theta_{{3, \text{dm}}}$ and $\sum_{i=1}^3{\bf \Delta}\theta_{{i, \text{dm}}}$. }
Fig.~\ref{fig8} displays several failure cases, {\em i.e.,} images produced by DMs with $\sum_{i=1}^3{\bf \Delta}\theta_{{i, \text{dm}}}$ but classified as forgotten concept categories.
\textcolor{blue!100}{Additionally, the results of SepME on  ${\bf \Delta}\theta_{{i, \text{dm}}}$ and $\sum_{i\in\{j,k\}}{\bf \Delta}\theta_{{i, \text{dm}}}$ indicate the feasibility of concept restoration after multi-concept erasures.}

Next, we individually fine-tune  ${\bf \Delta}\theta_{{1, \text{dm}}}$, ${\bf \Delta}\theta_{{2, \text{dm}}}$, and ${\bf \Delta}\theta_{{3, \text{dm}}}$ before combining them to simultaneously eliminate multiple concepts. The experimental results are detailed in Tab.~\ref{tab4}. It is apparent that all unlearning methods effectively erase forgotten concepts. However, both AbConcept and our G-CiRs exhibit shortcomings in restoring forgotten concepts. For example, Abconcept and G-CiRs under the setting $\theta_\text{dm}$+$\sum_{i\in\{1,3\}}{\bf \Delta}\theta_{{i, \text{dm}}}$ only perform 68\% and 71.6\% classification accuracy for `Van Gogh', respectively.
In contrast, our SepME$_1$ in Tab. \ref{tab4} demonstrates nearly perfect recovery of erased concepts. \textcolor{blue!100}{This emphasizes the effectiveness of our SepME in restoring forgotten concepts and the feasibility of separately optimizing ${\bf \Delta}\theta_{i,\text{dm}}$.}

{\bf Iterative-concept erasure.}
To realize multi-concept erasure and concept restoration under this scenario, we avoid sequentially fine-tuning model weights, as restoring the early weights ${\bf \Delta}\theta_{{i, \text{dm}}}$ inevitably affects the erasure performance of ${\bf \Delta}\theta_{{>i, \text{dm}}}$. Inspired by the success of previous experiments where we optimized ${\bf \Delta}\theta_{{1\sim 3, \text{dm}}}$ individually, we achieve iterative concept erasure through this optimization mode.

In the initial unlearning step, G-CiRs is employed to fine-tune all parameters of cross-attention modules in DMs. 
This enables better unlearning of the forgotten concept with smaller weight modifications. 
In each subsequent unlearning step $t$, we recalculate $S_p$ using $c_{<t, f}$ to construct the weight increments of to$\_$k and to$\_$v layers. These increments are further fine-tuned to erase $c_{t,f}$. The experimental results presented as SepME$_2$ in Tab.~\ref{tab4} show comparable performance to SepME$_1$. \textcolor{blue!100}{Overall, SepME can effectively achieve iterative concept erasure and concept restoration.}

\subsection{Object Removal.} The experimental results on object removal yield similar conclusions to those on style removal. For detailed information, please refer to the appendix.

\section{Conclusion}
In this study, we present an innovative machine unlearning technique for diffusion models, namely separable multi-concept eraser (SepME). SepME leverages a correlation term and momentum statistics to yield concept-irrelevant representations. It not only maintains overall model performance during concept erasure but also adeptly balances loss magnitudes across multiple concepts. Furthermore, SepME allows for the separation of weight increments, providing flexibility in manipulating various concepts, including concept restoration and iterative concept erasure. Extensive experiments validate the effectiveness of our methods. 

{\bf Broader Impact.}
As the field of deep learning continues to evolve, it presents both exciting opportunities and profound responsibilities for our community. While recent advances hold promise for solving complex problems, they also raise concerns regarding ethical and societal implications.
As researchers in this domain, we recognize our obligation to comprehend and address the challenges associated with the widespread adoption of deep learning technology. Machine learning models, despite their potential benefits, can harbor harmful biases, unintended behaviors, and pose risks to user privacy.
Our work contributes to this discourse by proposing a post-processing `unlearning' phase aimed at mitigating these concerns. Through extensive empirical investigation, we demonstrate progress over previous solutions in practical settings. However, it's important to acknowledge that while our approach, SpeME, represents a significant step forward, we cannot claim perfect mitigation of these issues.
Therefore, it's imperative that caution is exercised in the practical application of deep learning techniques, and that rigorous auditing and evaluation of machine learning models are conducted.


\bibliography{iccv2023AuthorKit/egbib}
\bibliographystyle{ieee_fullname}

\newpage
\appendix
\onecolumn

\section{Code availability}
The code is available at \url{https://github.com/Dlut-lab-zmn/SepCE4MU}.
\section{Algorithms}
The algorithmic details for generating concept-irrelevant representations are outlined in Alg. \ref{alg:algorithm1}. Furthermore, Alg. \ref{alg:algorithm2} provides a comprehensive explanation of the separable multi-concept eraser.
\begin{algorithm}[htpb]
	\caption{G-CiRs.}
	\label{alg:algorithm1}
	\KwIn{The diffuser $\text{G}(\cdot)$, the frozen weights $\theta_\text{dm}$, the weight increment ${\bf \Delta} \theta_\text{dm}$, the $\text{N}$ forgotten concepts $c_{i,f}\in c_f$,  the blank prompt $c_\emptyset$, the inference dataset $x_0\in D$, the noise schedule $\Bar{\alpha_t}$, the hyperparameter $\lambda$.}
	\KwOut{The fine-tuned model increment ${\bf \Delta} \theta_\text{dm}$.}  
	\BlankLine
	\For{$n$, $x_0\in D$}{
        Randomly select a sampling step $t$;
        
        $x_t = \sqrt{\Bar{\alpha_t}}x_0 + \sqrt{1-\Bar{\alpha_t}}\epsilon$, $\epsilon\in \mathcal{N}(0,\mathbf{I})$;

        $\epsilon_{c_{f}}$ = $\text{G}_{\theta_\text{dm}}(x_t, c_{f}, t)$;

        $\epsilon_{c_\emptyset}$ = $\text{G}_{\theta_\text{dm}}(x_t, c_\emptyset, t)$;

        $\epsilon_{c_{f}}^\prime$ = $\text{G}_{\theta_\text{dm} + {\bf \Delta} \theta_\text{dm}}(x_t, c_{f}, t)$;

        $\epsilon_{c_\emptyset}^\prime$ = $\text{G}_{\theta_\text{dm} + {\bf \Delta} \theta_\text{dm}}(x_t, c_\emptyset, t)$;

        $\mathcal{L}_\text{cor}(c_{f},{\bf \Delta}\theta_\text{dm}) = \text{Avg}((\epsilon_{c_{f}} - \epsilon_{c_\emptyset}) \cdot (\epsilon_{c_{f}}^\prime - \epsilon_{c_\emptyset}^\prime))$;
        
        $\eta_i$ = $\frac{\|\mathcal{L}_\text{cor}(c_{1,f},{\bf \Delta}\theta_\text{dm})\|_2}{\|\mathcal{L}_\text{cor}(c_{i,f},{\bf \Delta}\theta_\text{dm})\|_2}$;

        $\mathcal{L}_\text{mom}^n = \alpha\mathcal{L}_\text{mom}^{n-1} + (1-\alpha)\sum_{i=1}^\text{N}\eta_i\mathcal{L}_\text{cor}(c_{i,f},{\bf \Delta}\theta_\text{dm})$;

        \If{$\mathcal{L}_\text{mom}^n\leq \tau$}{
            break;
        }

        $\min_{{\bf \Delta}\theta_\text{dm}}\mathcal{L}_\text{G-CiRs}$ = $\sum_{i=1}^\text{N}\eta_i\mathcal{L}_\text{cor}(c_{i,f},{\bf \Delta}\theta_\text{dm}) + \lambda\|{\bf \Delta}\theta_\text{dm}\|_p$
        
	}
\end{algorithm}

\begin{algorithm}[t]
	\caption{SepME.}
	\label{alg:algorithm2}
	\KwIn{The diffuser $\text{G}(\cdot)$, the frozen weights $\theta_\text{dm}$, the weight increment ${\bf \Delta} \theta_\text{dm}$, the $\text{N}$ forgotten concepts $c_{i,f}\in c_f$,  the blank prompt $c_\emptyset$, the inference dataset $x_0\in D$, the noise schedule $\Bar{\alpha_t}$, the hyperparameters $\lambda$ and $\beta$.}
	\KwOut{The fine-tuned weight increments ${\bf \Delta} \theta_{\text{i}\in[1,\text{N}], \text{dm}}$.}  
	\BlankLine

    \For{$c_{i,f}\in c_f$}{
        $A$ = $[c_\emptyset^\top; c_{1,f}^\top; \cdots; c_{i-1,f}^\top; c_{i+1,f}^\top; \cdots; c_{\text{N},f}^\top]^\top$;
        
        Obtain solutions $S_p$, $A\otimes S_p = 0$;

        \For{ ${\bf \Delta}\theta_{\text{to}\_\text{k}}$ $\in$ ${\bf \Delta} \theta_\text{i, dm}$}{
            Initialize $w\in\mathcal{W}$ to a zero matrix;
            
            ${\bf \Delta}\theta_{\text{to}\_\text{k}} = (w \otimes (\beta S_p^\top))^\top$
        }
        \For{ ${\bf \Delta}\theta_{\text{to}\_\text{v}}$ $\in$ ${\bf \Delta} \theta_\text{i, dm}$}{
            Initialize $w\in\mathcal{W}$ to a zero matrix;
            
            ${\bf \Delta}\theta_{\text{to}\_\text{v}} = (w \otimes (\beta S_p^\top))^\top$
        }
    }
	\For{$n, x_0\in D$}{
        Randomly select a sampling step $t$;
        
        $x_t = \sqrt{\Bar{\alpha_t}}x_0 + \sqrt{1-\Bar{\alpha_t}}\epsilon$, $\epsilon\in \mathcal{N}(0,\mathbf{I})$;

        $\epsilon_{c_{f}}$ = $\text{G}_{\theta_\text{dm}}(x_t, c_{f}, t)$;

        $\epsilon_{c_\emptyset}$ = $\text{G}_{\theta_\text{dm}}(x_t, c_\emptyset, t)$;

        \For{$c_{i,f}\in c_{f}$}{
            $\epsilon_{c_{i, f}}^\prime$ = $\text{G}_{\theta_\text{i, dm} + {\bf \Delta} \theta_\text{dm}}(x_t, c_{i, f}, t)$;
    
            $\epsilon_{c_\emptyset}^\prime$ = $\text{G}_{\theta_\text{i, dm} + {\bf \Delta} \theta_\text{dm}}(x_t, c_\emptyset, t)$;

            $\mathcal{L}_\text{cor}(c_{i, f},{\bf \Delta}\theta_\text{dm}) = \text{Avg}((\epsilon_{c_{i, f}} - \epsilon_{c_\emptyset}) \cdot (\epsilon_{c_{i, f}}^\prime - \epsilon_{c_\emptyset}^\prime))$;
    
            $\eta_i$ = $\frac{\|\mathcal{L}_\text{cor}(c_{1,f},{\bf \Delta}\theta_\text{dm})\|_2}{\|\mathcal{L}_\text{cor}(c_{i,f},{\bf \Delta}\theta_\text{dm})\|_2}$;
    
        }

        $\mathcal{L}_\text{mom}^n = \alpha\mathcal{L}_\text{mom}^{n-1} + (1-\alpha)\sum_{i=1}^\text{N}\eta_i\mathcal{L}_\text{cor}(c_{i,f},{\bf \Delta}\theta_\text{dm})$;

        \If{$\mathcal{L}_\text{mom}^n\leq \tau$}{
            break;
        }

        $\min_\mathcal{W}\mathcal{L}_\text{SepME}$ = $\sum_{i=1}^\text{N}\eta_i\mathcal{L}_\text{cor}(c_{i,f},{\bf \Delta}\theta_{{i, \text{dm}}})$ + $\lambda\|\mathcal{W}\|_p$
        
	}
\end{algorithm}

\section{Style removal}
\begin{figure*}[t]
    \centering
    \includegraphics[width=0.75\linewidth]{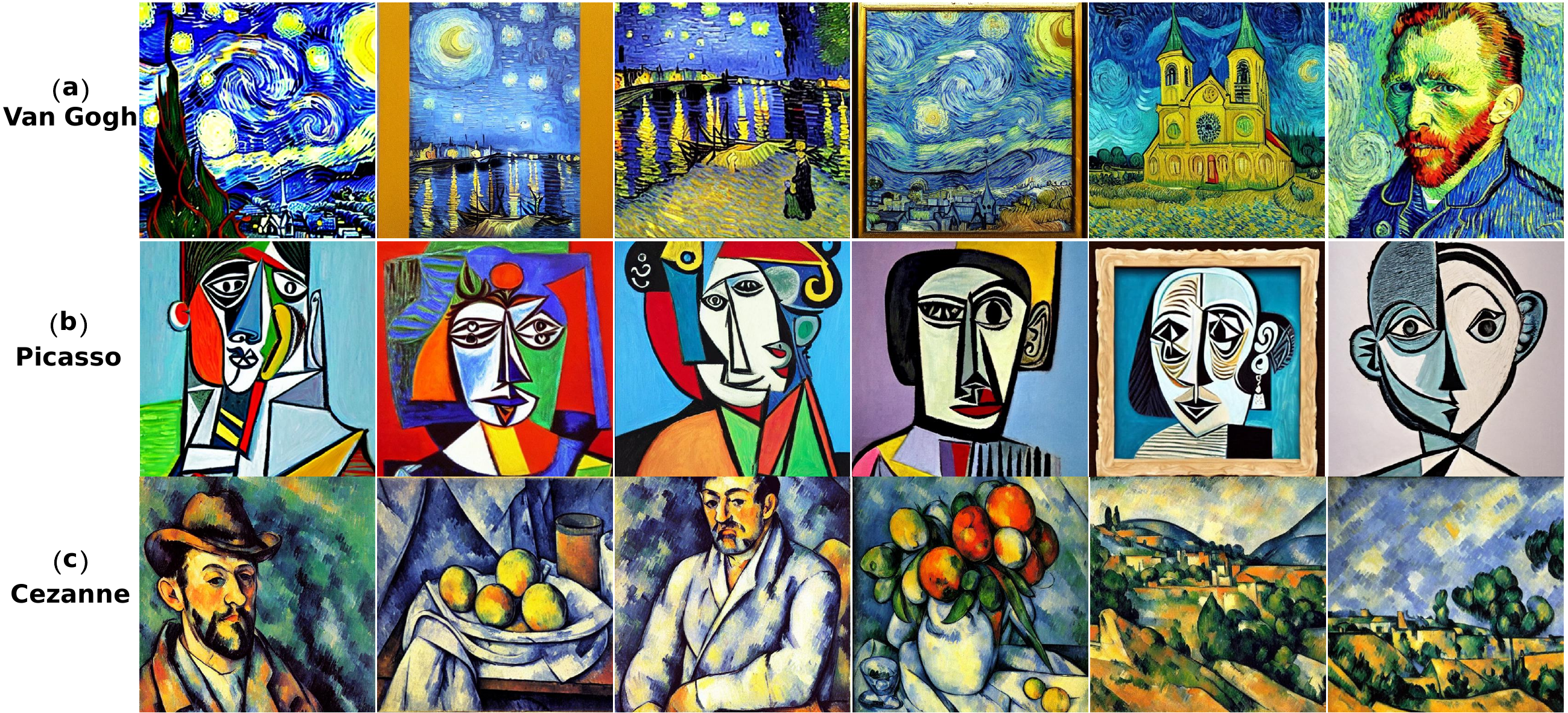}
    \caption{Visual examples produced by original diffusion models.}
    \label{fig:sup1}
\end{figure*}

\textit{Reference images $x_0$.} Several visual examples generated by the original diffusion model are provided in Fig. \ref{fig:sup1}.

\begin{figure*}[t]
    \centering
    \includegraphics[width=0.75\linewidth]{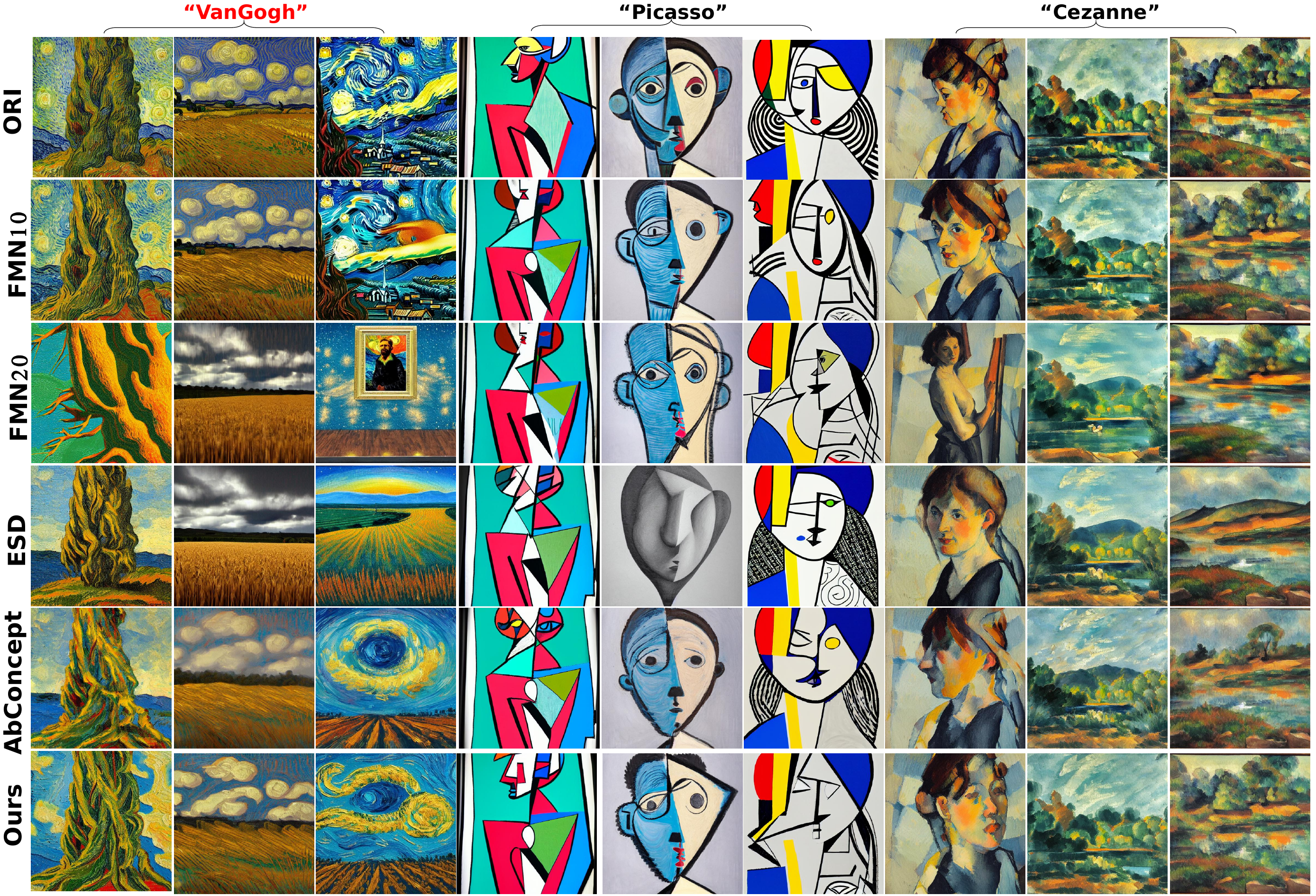}
    \caption{Qualitative comparison among various unlearning techniques for DMs with `Van Gogh' as the erased concept.}
    \label{fig:sup2}
\end{figure*}

\textit{Evaluation for G-CiRs.} Additional visual examples produced with sentences\footnote{\url{https://github.com/rohitgandikota/erasing/blob/main/data/art_prompts.csv}} containing artist names as prompts for various unlearned DMs are shown in Figs.~\ref{fig:sup2}$\sim$\ref{fig:sup4}.

\textit{Other experimental results.} 1) The more detailed results of Tab.~\ref{tab4} are shown in Tab.~\ref{suptab1}.
2) Furthermore, we perform an ablation study on the threshold $\tau$, which controls the moment of early stopping.
The detailed experimental results are shown in Tab.~\ref{suptab2}.
Observations reveal that G-CiRs achieves optimal erasing performance for `Van Gogh,' `Picasso,' and `Cezanne' at $\tau$ values of -5e-4, 0, and 0, respectively.

\section{Object Removal}
\textit{Evaluation Settings:} We employ the pre-trained ResNet50 \cite{he2016deep} as the object classification network. We analyze nine classes within Imagenette \cite{howard2020fastai}, excluding the `cassette player' category due to ResNet50's classification accuracy falling below 50\% on this class. During evaluation, we generate 250 images per class, {\em i.e.,} 50 seeds for each concept, and 5 images for each seed. 

\subsection{Evaluation for G-CiRs} 
For the \textit{single concept erasure}, one category is chosen as the forgotten concept, while others serve as evaluation concepts.
The quantitative results are presented in Tab.~\ref{suptab3}.
On one hand, our G-CiRs achieves optimal performance in terms of ACC metric across nine object categories. On the other hand, the proposed G-CiRs induces fewer modifications to model weights.
Additionally, the qualitative results are presented in Figs.~\ref{fig:sup5}$\sim$\ref{fig:sup7}.

\subsection{Evaluation for SepME} 

\textit{Separate optimization.}
Next, we individually train  ${\bf \Delta}\theta_{{1, \text{dm}}}$, ${\bf \Delta}\theta_{{2, \text{dm}}}$, and ${\bf \Delta}\theta_{{3, \text{dm}}}$ before combining them to simultaneously erase multiple concepts. The experimental results are detailed in Tab.~\ref{suptab4}. It is apparent that all unlearning methods effectively erase forgotten concepts. However, AbConcept exhibits shortcomings in restoring a forgotten concept. For example, Abconcept under the setting $\theta_\text{dm}$+$\sum_{i=2}^3{\bf \Delta}\theta_{{i, \text{dm}}}$ only perform 74\%classification accuracy for `chain saw', respectively.
In contrast, our SepME$_1$ in Tab. \ref{suptab4} demonstrates nearly perfect recovery of erased concepts. This emphasizes the effectiveness of our SepME in restoring forgotten concepts and the feasibility of separately optimizing ${\bf \Delta}\theta_{i,\text{dm}}$.

\textit{Iterative-concept erasure.}
The iterative concept erasure implies that each erasure step $t$ can only utilize knowledge of the previously forgotten concepts $c_{<t, f}$. 
To realize this erasure, we avoid sequentially fine-tuning model weights, as restoring the early weights ${\bf \Delta}\theta_{{i, \text{dm}}}$ inevitably affects the erasure performance of ${\bf \Delta}\theta_{{>i, \text{dm}}}$. Inspired by the success of separately optimizing ${\bf \Delta}\theta_{{1\sim 3, \text{dm}}}$, we endeavor to implement iterative concept erasure through this setting.
In the initial unlearning step, we employ G-CiRs to fine-tune all parameters of cross-attention modules in DMs. This enables better unlearning of the forgotten concept with fewer weight modifications. In each subsequent unlearning step $t$, we recalculate $S_p$ using $c_{<t, f}$ to construct the weight increments of to$\_$k and to$\_$v layers and fine-tune these increments to erase the concept $c_{t,f}$. The experimental results presented as SepME$_2$ in Tab.~\ref{suptab4} demonstrate comparable performance to SepME$_1$. Overall, SepME can effectively achieve iterative concept erasure and concept restoration.

\section{Other details.}
We omit the consideration of iterative erasure, where multiple concepts are erased at each step, as the weight increments for erasing various concept can be optimized separately.

\textbf{Cosine Function:} We explored the use of the cosine function as an alternative to $\mathcal{L}_{cor}$ and assessed its performance across various hyperparameters and learning rates. However, this approach did not yield satisfactory results. We attribute this to the significant prediction gap between model samples, {\em i.e.,} the normalization of sample constraints affects the optimization direction.
\begin{figure*}[t]
    \centering
    \includegraphics[width=0.8\linewidth]{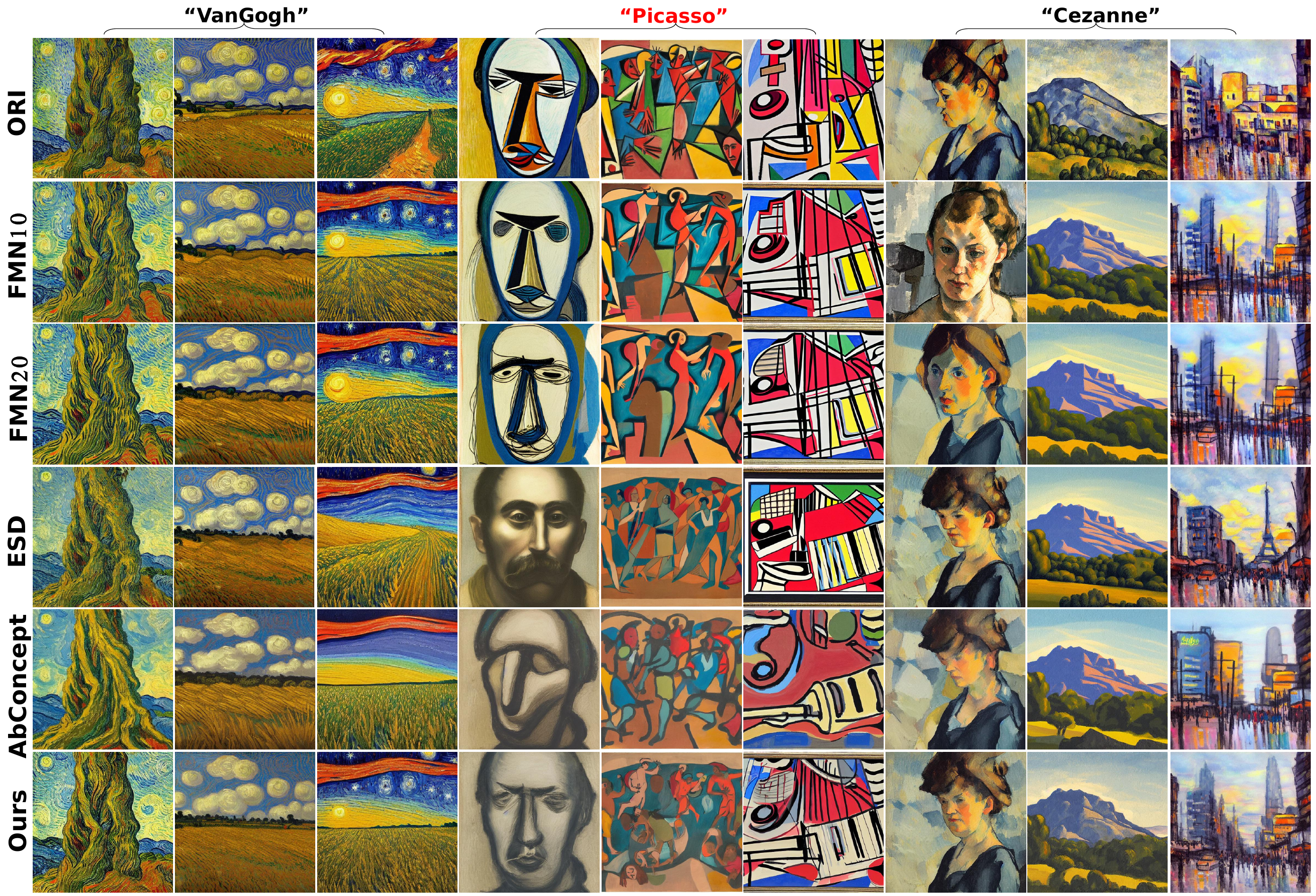}
    \caption{Qualitative comparison among various unlearning techniques for DMs with `Picasso' as the erased concept.}
    \label{fig:sup3}
\end{figure*}

\begin{figure*}[t]
    \centering
    \includegraphics[width=0.8\linewidth]{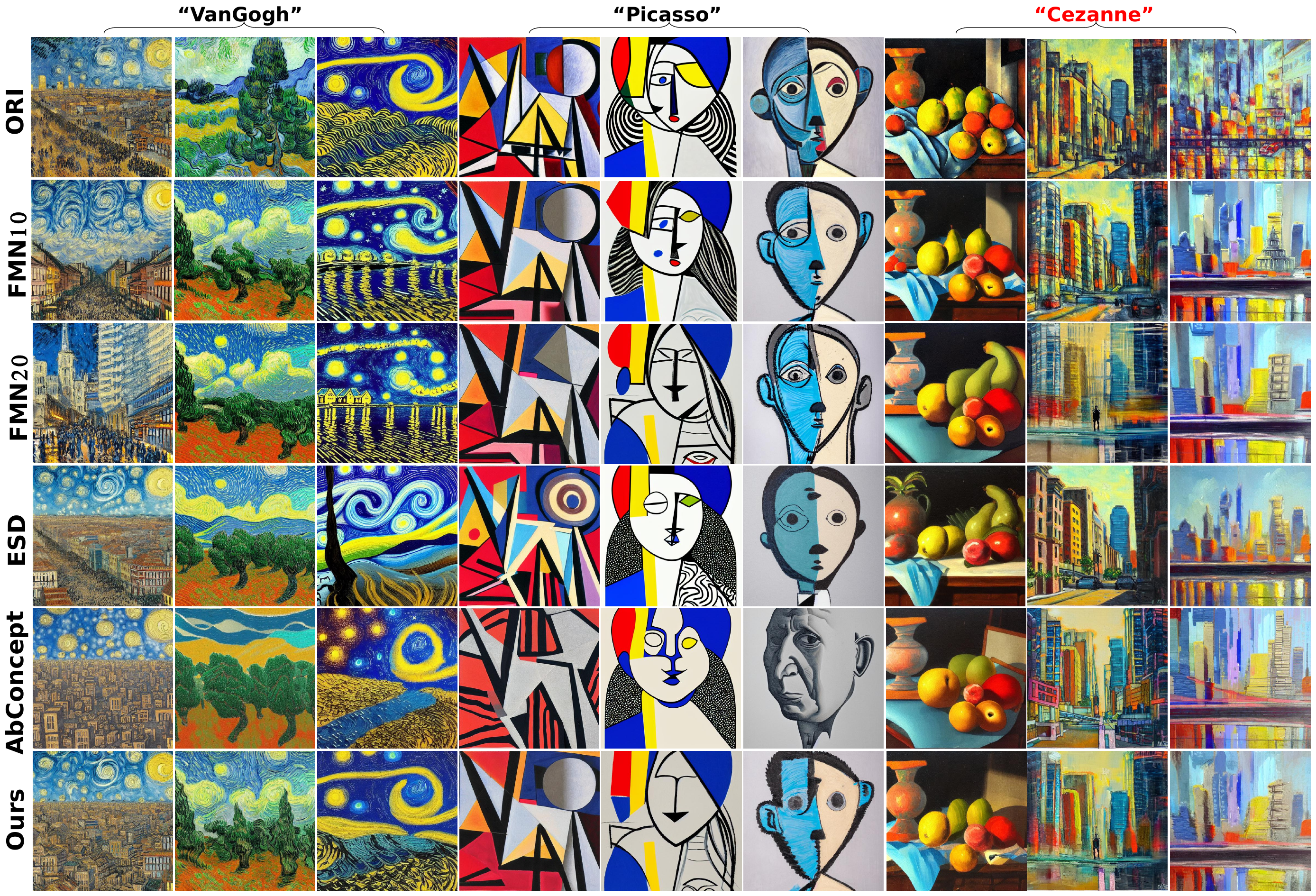}
    \caption{Qualitative comparison among various unlearning techniques for DMs with `Cezanne' as the erased concept.}
    \label{fig:sup4}
\end{figure*}

\begin{table*}[t]
    \tabcolsep = 0.1cm
    \begin{center}
    \caption{Quantitative results of SepME when separately fine-tuning ${\bf \Delta}\theta_{{1\sim 3, \text{dm}}}$. The concepts $c_\text{1,f}$, $c_\text{2,f}$, and $c_\text{3,f}$ correspond to `Van Gogh', `Cezanne', and `Picasso', respectively. In SepME$_1$, ${\bf \Delta}\theta_{{1\sim 3, \text{dm}}}$ are separately optimized when all forgotten concepts are known. SepME$_2$ follows the mode of iterative concept erasure. In other words, the $t$-th erasure step only possesses knowledge ($Kn$) of the previously forgotten concepts. `xattn' means that we employ all layers of cross attention modules.
    }
    \label{suptab1}
    \begin{tabular}{c|cc ccc cc}
    \hline
    &\multicolumn{7}{c}{SepME$_1$--({ACC/LPIPS})--$Kn = [c_\text{1,f}; c_\text{2,f}; c_\text{3,f}]$}\\
    $\theta_{{\text{dm}}}$&+${\bf \Delta}\theta_{{1, \text{dm}}}$&+${\bf \Delta}\theta_{{2, \text{dm}}}$&+${\bf \Delta}\theta_{{3, \text{dm}}}$&+$\sum_{i=1}^2{\bf \Delta}\theta_{{i, \text{dm}}}$&+$\sum_{i\in\{1,3\}}{\bf \Delta}\theta_{{i, \text{dm}}}$&+$\sum_{i=2}^3{\bf \Delta}\theta_{{i, \text{dm}}}$&+$\sum_{i=1}^3{\bf \Delta}\theta_{{i, \text{dm}}}$\\
    VG&\textcolor{gray!80}{0.356/0.364}&\textcolor{gray!80}{1.000/0.182}&\textcolor{gray!80}{0.908/0.182}&0.308/0.365&0.308/0.365&\underline{0.956}/0.181&{0.308/0.364}\\
    CE&\textcolor{gray!80}{1.000/0.144}&\textcolor{gray!80}{0.320/0.304}&\textcolor{gray!80}{1.000/0.144}&0.320/0.304&\underline{1.000}/0.144&0.288/0.304&{0.320/0.304}\\
    PC&\textcolor{gray!80}{1.000/0.185}&\textcolor{gray!80}{1.000/0.185}&\textcolor{gray!80}{0.000/0.460}&\underline{1.000}/0.185&0.000/0.460&0.000/0.460&{0.000/0.460}\\
    Others&\textcolor{gray!80}{0.957/0.194}&\textcolor{gray!80}{0.891/0.243}&\textcolor{gray!80}{0.955/0.184}&0.855/0.253&0.957/0.211&0.845/0.248&{\underline{0.803}/0.260}\\
    \hline
    &\multicolumn{7}{c}{SepME$_2$--({ACC/LPIPS})}\\
    $Kn$&$[c_\text{1,f}]$&$[c_\text{1,f};c_\text{2,f}]$&$[c_\text{1,f};c_\text{2,f};c_\text{3,f}]$&-&-&-&-\\
    VG&\textcolor{gray!80}{0.228/0.363}&\textcolor{gray!80}{0.956/0.182}&\textcolor{gray!80}{0.956/0.182}&0.228/0.363&0.228/0.363&\underline{0.956}/0.182&{0.228/0.363}\\
    CE&\textcolor{gray!80}{1.000/0.182}&\textcolor{gray!80}{0.000/0.404}&\textcolor{gray!80}{1.000/0.172}&0.000/0.376&\underline{1.000}/0.177&0.000/0.411&{0.052/0.394}\\
    PC&\textcolor{gray!80}{0.964/0.150}&\textcolor{gray!80}{1.000/0.144}&\textcolor{gray!80}{0.288/0.277}&\underline{0.964}/0.150&0.108/0.292&0.288/0.277&{0.108/0.292}\\
    Others&\textcolor{gray!80}{0.994/0.173}&\textcolor{gray!80}{0.925/0.196}&\textcolor{gray!80}{0.986/0.185}&0.946/0.211&0.970/0.194&0.924/0.231&{\underline{0.916}/0.235}\\
    \hline
    &\multicolumn{7}{c}{Abconcept--({ACC/LPIPS})--xattn}\\
    VG&\textcolor{gray!80}{0.000/0.469}&\textcolor{gray!80}{0.728/0.257}&\textcolor{gray!80}{0.684/0.227}&0.000/0.472&0.000/0.469&\underline{0.636}/0.277&{0.000/0.487}\\
    CE&\textcolor{gray!80}{0.820/0.217}&\textcolor{gray!80}{0.036/0.360}&\textcolor{gray!80}{0.752/0.194}&0.288/0.299&\underline{0.680}/0.234&0.252/0.288&{0.144/0.349}\\
    PC&\textcolor{gray!80}{0.900/0.190}&\textcolor{gray!80}{0.852/0.251}&\textcolor{gray!80}{0.000/0.397}&\underline{0.600}/0.277&0.152/0.304&0.200/0.334&{0.052/0.364}\\
    Others&\textcolor{gray!80}{0.952/0.185}&\textcolor{gray!80}{0.923/0.215}&\textcolor{gray!80}{0.971/0.173}&0.861/0.227&0.891/0.194&0.903/0.206&{\underline{0.783}/0.247}\\
    \hline
    &\multicolumn{7}{c}{Abconcept--({ACC/LPIPS})--(to$\_$k;to$\_$v)}\\
    VG&\textcolor{gray!80}{0.592/0.329}&\textcolor{gray!80}{0.820/0.215}&\textcolor{gray!80}{0.728/0.271}&0.320/0.383&0.184/0.378&\underline{0.272}/0.398&{0.184/0.411}\\
    CE&\textcolor{gray!80}{1.000/0.193}&\textcolor{gray!80}{0.200/0.266}&\textcolor{gray!80}{0.800/0.222}&0.252/0.274&\underline{0.600}/0.230&0.052/0.307&{0.000/0.311}\\
    PC&\textcolor{gray!80}{1.000/0.142}&\textcolor{gray!80}{0.964/0.141}&\textcolor{gray!80}{0.216/0.306}&\underline{0.856}/0.147&0.320/0.316&0.072/0.347&{0.144/0.356}\\
    Others&\textcolor{gray!80}{0.973/0.178}&\textcolor{gray!80}{0.917/0.208}&\textcolor{gray!80}{0.952/0.191}&0.889/0.223&0.937/0.197&0.860/0.221&\underline{0.805}/0.248\\
    \hline
    &\multicolumn{7}{c}{G-CiRs--({ACC/LPIPS})--xattn}\\
    VG&\textcolor{gray!80}{0.228/0.363}&\textcolor{gray!80}{0.908/0.180}&\textcolor{gray!80}{0.908/0.276}&0.184/0.386&0.092/0.382&\underline{0.544}/0.276&{0.184/0.410}\\
    CE&\textcolor{gray!80}{1.000/0.154}&\textcolor{gray!80}{0.000/0.363}&\textcolor{gray!80}{1.000/0.189}&0.108/0.261&\underline{0.716}/0.178&0.216/0.313&{0.036/0.331}\\
    PC&\textcolor{gray!80}{1.000/0.175}&\textcolor{gray!80}{1.000/0.204}&\textcolor{gray!80}{0.052/0.358}&\underline{1.000}/0.220&0.452/0.247&0.200/0.280&{0.200/0.303}\\
    Others&\textcolor{gray!80}{0.994/0.173}&\textcolor{gray!80}{0.912/0.222}&\textcolor{gray!80}{0.950/0.179}&0.940/0.202&0.969/0.195&0.890/0.224&\underline{0.843}/0.268\\
    \hline
    &\multicolumn{7}{c}{G-CiRs--({ACC/LPIPS})--(to$\_$k;to$\_$v)}\\
    VG&\textcolor{gray!80}{0.344/0.356}&\textcolor{gray!80}{0.908/0.209}&\textcolor{gray!80}{0.820/0.236}&0.228/0.378&0.228/0.381&\underline{0.320}/0.344&{0.136/0.397}\\
    CE&\textcolor{gray!80}{0.892/0.155}&\textcolor{gray!80}{0.356/0.263}&\textcolor{gray!80}{0.752/0.219}&0.152/0.282&\underline{0.900}/0.210&0.152/0.295&{0.152/0.334}\\
    PC&\textcolor{gray!80}{1.000/0.184}&\textcolor{gray!80}{1.000/0.192}&\textcolor{gray!80}{0.148/0.290}&\underline{0.832}/0.164&0.252/0.293&0.144/0.319&{0.072/0.346}\\
    Others&\textcolor{gray!80}{0.944/0.193}&\textcolor{gray!80}{0.908/0.212}&\textcolor{gray!80}{0.920/0.203}&{0.873}/0.237&0.938/0.220&0.848/0.258&\underline{0.791}/0.272\\
    \hline
    \end{tabular}
    \end{center}
    \vspace{-7mm}
\end{table*}

\begin{table*}[t]
    \tabcolsep = 0.1cm
    \begin{center}
    \caption{Ablation study to investigate the influence of the hyperparameter $\tau$ on unlearning performance.
    }
    \label{suptab2}
    \begin{tabular}{c|cc ccc cc}
    \hline
Erased (G-CiRs)& $\tau$ & Van Gogh & Picasso & Cezanne & Others &$\|{\bf \Delta}\theta_\text{dm}\|_p\downarrow$\\
\hline
\multirow{5}{*}{Van Gogh}&
1e-3 & 0.684/0.310 & 1.000/0.180 & 1.000/0.140 & 1.000/0.159 & 24.28\\
&5e-4 & 0.592/0.334 &1.000/0.178 & 1.000/0.140 & 1.000/0.159 & 27.43\\
&0. & 0.456/0.353 & 1.000/0.171 & 1.000/0.145 & 1.000/0.166 & 36.24\\
&-5e-4 & 0.228/0.363 & 1.000/0.175 & 1.000/0.154 & 0.992/0.173 & 43.17\\
&-1e-3 & 0.092/ 0.419 & 1.000/ 0.195 & 0.892/0.153 & 0.976/0.176 & 51.63\\
\hline
\multirow{5}{*}{Picasso}
&5e-4 & 0.956/0.173 & 1.000/0.181 & 1.000/ 0.144 & 1.000/0.154 & 1.522\\
&5e-5 & 0.924/0.239 & 0.084/0.319 & 1.000/ 0.173 & 0.976/0.161 & 27.58\\
&0. & 0.908/0.276 & 0.052/0.358 & 1.000/ 0.189 & 0.952/0.179 & 36.90\\
&-5e-4 & 0.044/0.573 & 0.000/0.567 & 0.000/0.426 & 0.812/0.325 & 65.03\\
&-1e-3 & 0.000/0.604 & 0.000/0.578 & 0.036/0.516 & 0.696/0.379 & 67.89\\
\hline
\multirow{5}{*}{Cezanne}&
1e-4 & 0.772/0.216 &1.000/0.182 & 0.276/0.071 & 0.976/0.179 & 23.51\\
&5e-5 & 0.772/0.242 &1.000/0.215 & 0.072/0.355 & 0.960/0.195 & 30.67\\
&0.& 0.908/0.180 & 1.000/0.204 & 0.000/0.363& 0.912/0.222 & 36.55\\
&-5e-5 & 0.636/0.398 & 0.852/0.228 & 0.000/0.495 & 0.904/0.247 & 41.20\\
&-1e-4 & 0.272/0.458 & 0.852/0.251 & 0.000/0.543 & 0.832/0.281 & 45.75\\
\hline
\end{tabular}
\end{center}
\vspace{-7mm}
\end{table*}

\begin{table*}[t]
    \tabcolsep = 0.07cm
    \begin{center}
    \caption{Quantitative results of the single concept erasure. 
    $i$ in FMN$_i$ represents the iteration step.
    }
    \label{suptab3}
    \begin{tabular}{l|cccc ccccc}
    \hline
\multirow{2}{*}{ACC/LPIPS}&ORI&\multicolumn{2}{c}{FMN$_{20}$}&\multicolumn{2}{c}{FMN$_{50}$}&\multicolumn{2}{c}{AbConcept}&\multicolumn{2}{c}{G-CiRs}\\
&&Erased&Others&Erased&Others&Erased&Others&Erased&Others\\
\hline
chain saw &0.96/0.&0.84/0.241  &0.903/0.168&0.00/0.420   &0.773/0.269&0.28/0.325  &0.833/0.188&0.18/0.311  &0.825/0.214\\
church&0.84/0.&0.84/0.243  &0.893/0.167&0.06/0.440   &0.870/0.187&0.30/0.345  &0.870/0.183&0.16/0.255  &0.863/0.184\\
gas pump&0.80/0.&0.20/0.256  &0.930/0.170&0.02/0.379   &0.835/0.263&0.20/0.359  &0.905/0.186&0.12/0.341  &0.893/0.178\\
tench&0.88/0.&0.26/0.412   &0.873/0.178&0.00/0.462   &0.818/0.256&0.08/0.403  &0.870/0.173&0.08/0.381  &0.878/0.177\\
garbage truck&0.94/0.&0.48/0.229  &0.875/0.175&0.04/0.481   &0.735/0.314&0.58/0.293  &0.820/0.199&0.40/0.339  &0.815/0.213\\
english springer&1.00/0.&0.82/0.221  &0.873/0.167&0.02/0.364   &0.810/0.256&0.14/0.260  &0.873/0.232&0.00/0.292  &0.833/0.232\\
golf ball&1.00/0.&0.86/0.277  &0.850/0.171&0.44/0.420   &0.765/0.222&0.00/0.459  &0.845/0.231&0.32/0.345  &0.875/0.218\\
parachute&0.98/0.&0.54/0.429  &0.858/0.172&0.02/0.493   &0.770/0.258&0.40/0.436  &0.853/0.202&0.28/0.448  &0.823/0.204\\
french horn&1.00/0.&0.22/0.377  &0.853/0.185&0.04/0.429   &0.665/0.291&0.00/0.435  &0.823/0.185&0.00/0.452  &0.818/0.188\\
\hline
average&0.93/0.&0.63/0.298&0.879/0.173 &0.07/0.432 &0.782/0.257&\textcolor{blue}{0.22}/0.368&  \textcolor{red}{0.855}/0.199& \textcolor{red}{0.17}/0.352&\textcolor{blue}{0.847}/0.201\\
\hline
$\|{\bf \Delta}\theta_\text{dm}\|_p\downarrow$&-&\multicolumn{2}{c}{\textcolor{blue}{89.30}}&\multicolumn{2}{c}{254.9}&\multicolumn{2}{c}{150.6}&\multicolumn{2}{c}{\textcolor{red}{37.92}}\\
\hline
\end{tabular}
\end{center}
\vspace{-7mm}
\end{table*}

\begin{table*}[t]
    \tabcolsep = 0.1cm
    \begin{center}
    \caption{Quantitative results of SepME when separately fine-tuning ${\bf \Delta}\theta_{{1\sim 3, \text{dm}}}$. ${\bf \Delta}\theta_{{1, \text{dm}}}$, ${\bf \Delta}\theta_{{2, \text{dm}}}$, ${\bf \Delta}\theta_{{3, \text{dm}}}$ are optimizable weights for erasing `chain saw', `gas pump' and `garbage truck', respectively. In SepME$_1$, ${\bf \Delta}\theta_{{1\sim 3, \text{dm}}}$ are separately optimized when all forgotten concepts are known. 
    SepME$_2$ follows the mode of iterative concept erasure. In other words, the $t$-th erasure step only possesses knowledge of the previously forgotten concepts. `xattn' means that we employ all layers of cross attention modules.
    }
    \label{suptab4}
    \begin{tabular}{c|ccc cccc}
    \hline
ACC/LPIPS&\multicolumn{7}{c}{SepME$_1$ (to$\_$k,to$\_$v)}\\
$\theta_\text{dm}$& +$\Delta \theta_{1,\text{dm}}$&+$\Delta \theta_{2,\text{dm}}$&+$\Delta \theta_{3,\text{dm}}$&+$\sum_{i=1}^2\Delta\theta_{i,\text{dm}}$&+$\sum_{i\in {1,3}}\Delta\theta_{i,\text{dm}}$&+$\sum_{i=2}^3\Delta\theta_{i,\text{dm}}$&+$\sum_{i=1}^3\Delta\theta_{i,\text{dm}}$\\
chain saw&0.080/0.311&0.940/0.177&0.940/0.182&0.140/0.314&0.120/0.312&0.960/0.184&0.120/0.314\\
gas pump&0.680/0.142&0.300/0.308&0.680/0.142&0.260/0.307&0.680/0.142&0.360/0.298&0.340/0.298\\
garbage truck&0.920/0.157&0.920/0.157&0.000/0.410&0.920/0.157&0.000/0.411&0.000/0.411&0.000/0.411\\
Others&0.900/0.194&0.932/0.166&0.892/0.208&0.912/0.196&0.840/0.243&0.863/0.221&0.833/0.247\\
\hline
&\multicolumn{7}{c}{AbConcept (xattn)}\\
chain saw&0.280/0.325&0.940/0.192&0.920/0.203&0.340/0.328&0.220/0.344&0.740/0.222&0.080/0.359\\
gas pump&0.560/0.175&0.200/0.359&0.520/0.182&0.520/0.310&0.460/0.212&0.560/0.299&0.400/0.331\\
garbage truck&0.900/0.180&0.860/0.171&0.580/0.293&0.800/0.202&0.620/0.238&0.720/0.221&0.580/0.264\\
Others&0.867/0.192&0.907/0.187&0.853/0.201&0.843/0.234&0.827/0.246&0.843/0.235&0.687/0.309\\
\hline
&\multicolumn{7}{c}{SepME$_2$ (to$\_$k,to$\_$v)}\\
chain saw&0.220/0.311 &0.940/0.176&0.940/0.181&0.220/0.311&0.120/0.321&0.980/0.181&0.140/0.333\\
gas pump& 0.540/0.187 & 0.380/0.214& 0.680/0.151&0.260/0.230& 0.580/0.187 & 0.480/0.200 & 0.280/0.224\\
garbage truck& 0.840/0.203 & 0.800/0.162& 0.000/0.410&0.660/0.185& 0.000/0.406& 0.000/0.396& 0.000/0.387\\
Others& 0.870/0.234 & 0.910/0.174 & 0.890/0.208 & 0.847/0.239& 0.743/0.275& 0.883/0.221& 0.710/0.278\\
\hline
\end{tabular}
\end{center}
\end{table*}

\begin{figure*}[t]
    \centering
    \includegraphics[width=0.9\linewidth]{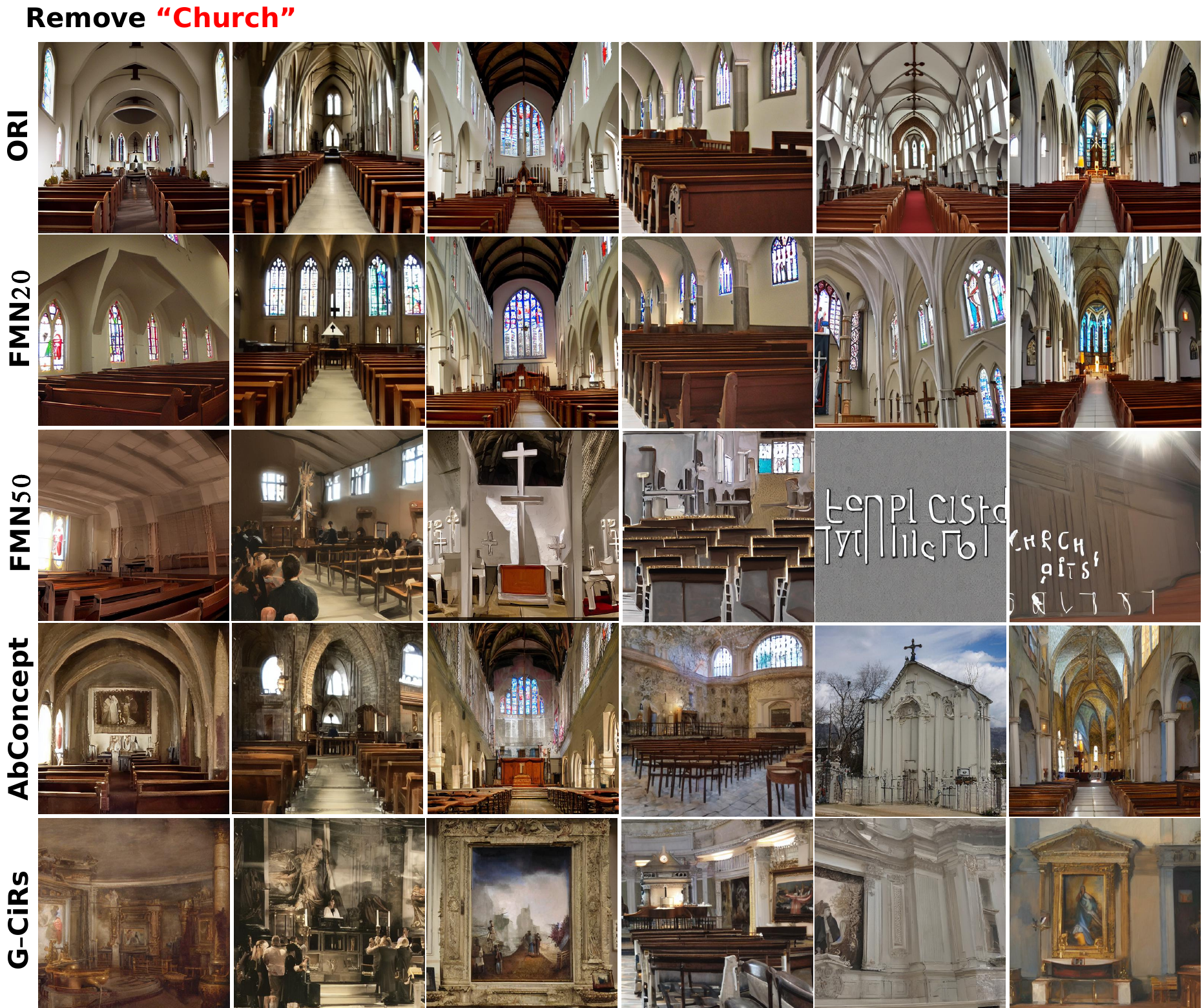}
    \caption{Qualitative comparison among various unlearning techniques for DMs with `church' as the erased concept.}
    \label{fig:sup5}
\end{figure*}
\begin{figure*}[t]
    \centering
    \includegraphics[width=0.9\linewidth]{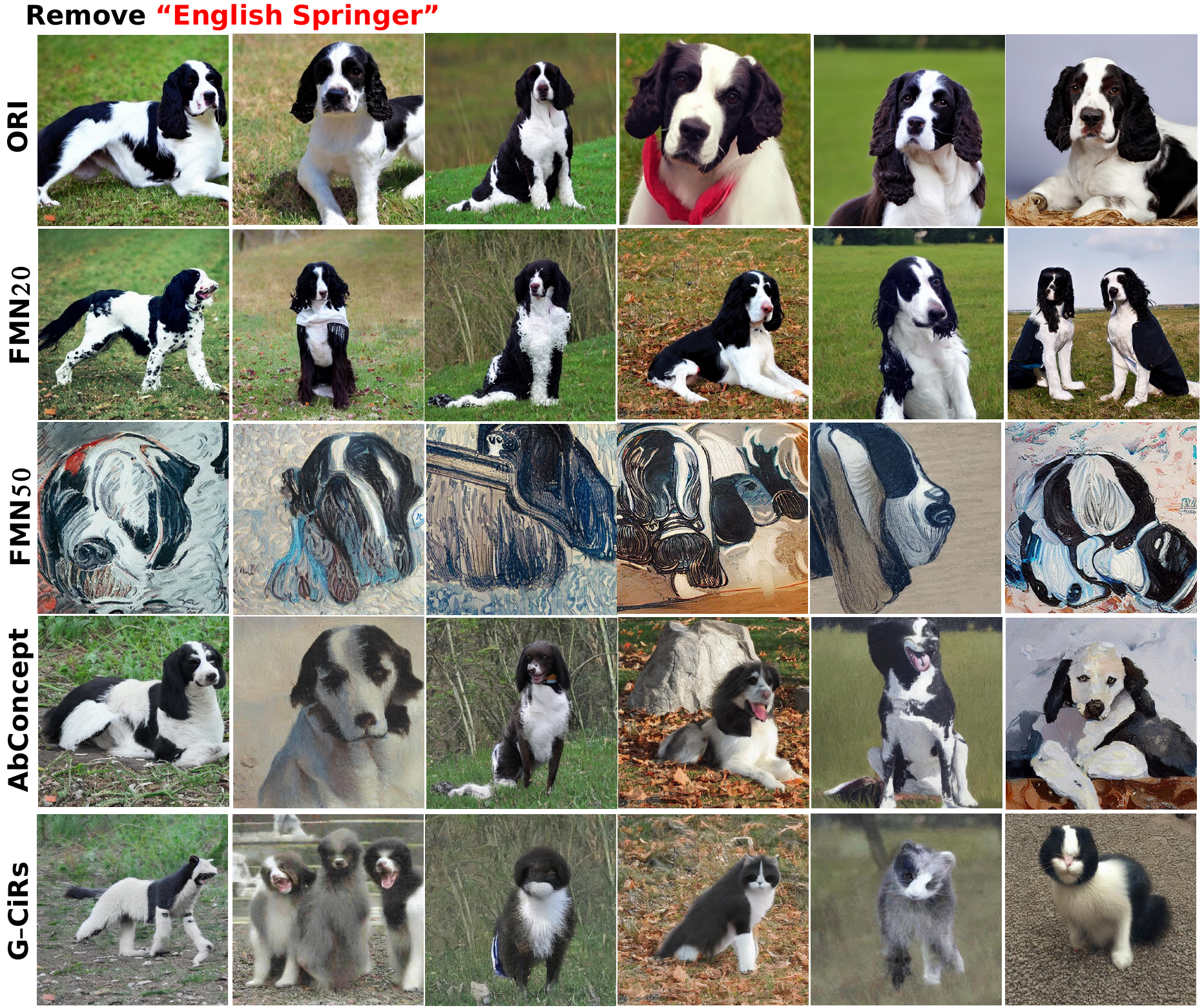}
    \caption{Qualitative comparison among various unlearning techniques for DMs with `English springer' as the erased concept.}
    \label{fig:sup6}
\end{figure*}
\begin{figure*}[t]
    \centering
    \includegraphics[width=0.9\linewidth]{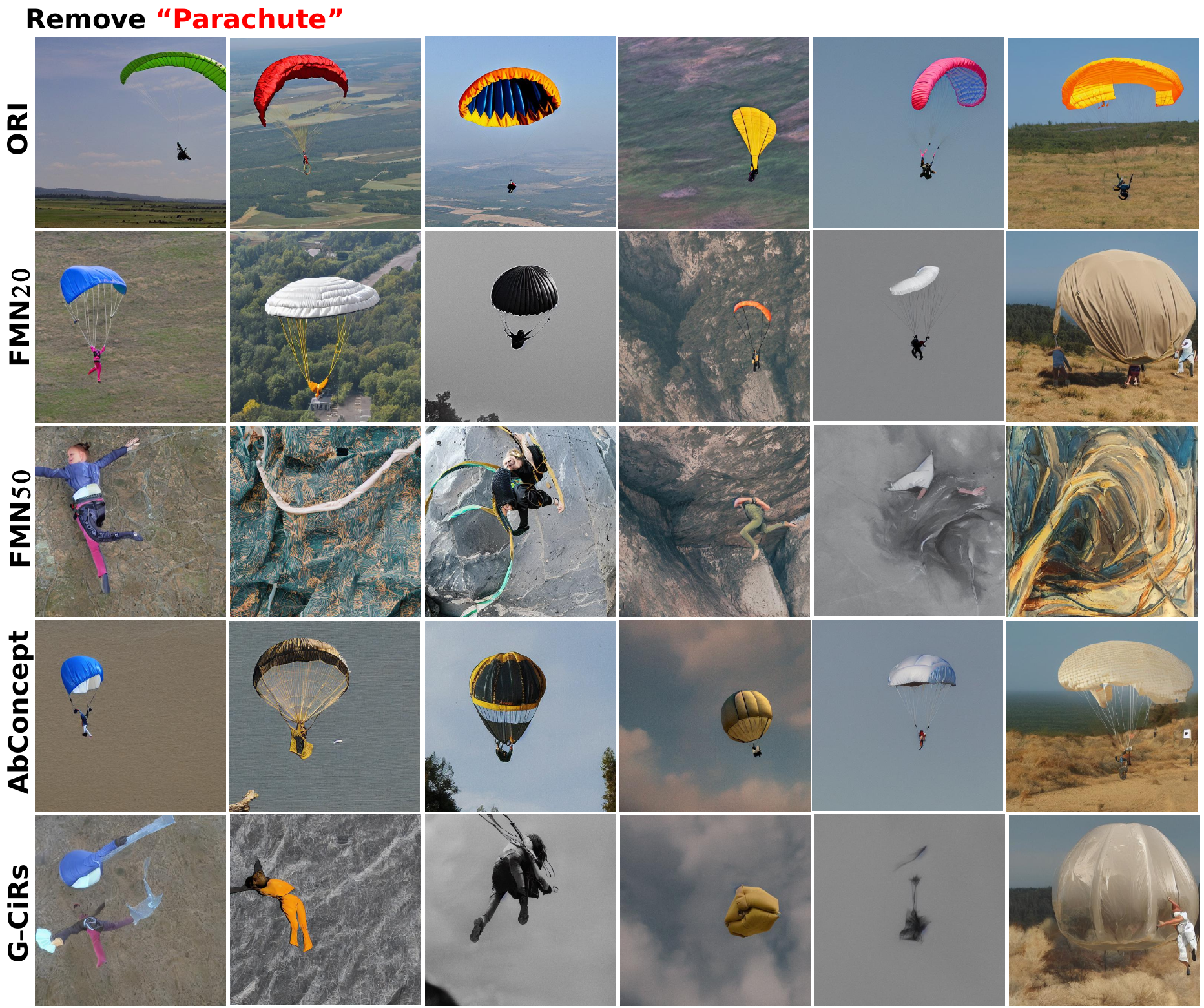}
    \caption{Qualitative comparison among various unlearning techniques for DMs with `parachute' as the erased concept.}
    \label{fig:sup7}
\end{figure*}

\end{document}